\documentclass[10pt]{article} 
\usepackage[preprint]{tmlr}

\usepackage{amsmath,amsfonts,bm}

\def\eqref#1{equation~\ref{#1}}

\def\1{\bm{1}}

\def\ervc{{\textnormal{c}}}

\def\ervx{{\textnormal{x}}}

\def\rmC{{\mathbf{C}}}

\def\rmX{{\mathbf{X}}}

\def\evx{{x}}

\def\mC{{\bm{C}}}

\DeclareMathAlphabet{\mathsfit}{\encodingdefault}{\sfdefault}{m}{sl}
\SetMathAlphabet{\mathsfit}{bold}{\encodingdefault}{\sfdefault}{bx}{n}

\usepackage{hyperref}
\usepackage{url}
\usepackage{graphicx}
\usepackage{booktabs}
\usepackage{multirow}
\usepackage{comment}

\title{Harnessing Large Language Models Over Transformer \\Models for Detecting Bengali Depressive Social Media Text: \\A Comprehensive Study}

\author{\name Ahmadul Karim Chowdhury \email ahmadulkc@gmail.com \\
      \addr Department of Computer Science and Engineering \\
      Ahsanullah University of Science and Technology
      \AND
      \name Md. Saidur Rahman Sujon \email sr.sujon.cyb@gmail.com \\
      \addr Department of Computer Science and Engineering
      \\
      Ahsanullah University of Science and Technology
      \AND
      \name Md. Shirajus Salekin Shafi \email salekin68@gmail.com
      \\
      \addr Department of Computer Science and Engineering\\
      Ahsanullah University of Science and Technology
      \AND
      \name Tasin Ahmmad \email tasinahmed411@gmail.com \\
      \addr Department of Computer Science and Engineering
      \\
      Ahsanullah University of Science and Technology
      \AND
      \name Sifat Ahmed \email sifat.austech@outlook.com \\
      \addr Co-founder \& Chief Technology Officer at Dolami Inc.
      \AND
      \name Khan Md Hasib \email khanmdhasib.aust@gmail.com \\
      \addr Assistant Professor, Department of Computer Science \& Engineering
      \\
      Bangladesh University of Business and Technology
      \AND
      \name Faisal Muhammad Shah \email 	faisal.cse@aust.edu \\
      \addr Associate Professor, Department of Computer Science and Engineering
      \\
      Ahsanullah University of Science and Technology}

\def\month{MM}  
\def\year{YYYY} 
\def\openreview{\url{https://openreview.net/forum?id=XXXX}} 

\begin{document}

\maketitle

\begin{abstract}
In an era where the silent struggle of underdiagnosed depression pervades globally, our research delves into the crucial link between mental health and social media. This work focuses on early detection of depression, particularly in extroverted social media users, using LLMs such as GPT 3.5, GPT 4 and our proposed  GPT 3.5 fine-tuned model DepGPT, as well as advanced Deep learning models(LSTM, Bi-LSTM, GRU, BiGRU) and Transformer models(BERT, BanglaBERT, SahajBERT, BanglaBERT-Base). The study categorized Reddit and X datasets into "Depressive" and "Non-Depressive" segments, translated into Bengali by native speakers with expertise in mental health, resulting in the creation of the Bengali Social Media Depressive Dataset (BSMDD). Our work provides full architecture details for each model and a methodical way to assess their performance in Bengali depressive text categorization using zero-shot and few-shot learning techniques. Our work demonstrates the superiority of SahajBERT and Bi-LSTM with FastText embeddings in their respective domains also tackles explainability issues with transformer models and emphasizes the effectiveness of LLMs, especially DepGPT, demonstrating flexibility and competence in a range of learning contexts. According to the experiment results, the proposed model, DepGPT, outperformed not only Alpaca Lora 7B in zero-shot and few-shot scenarios but also every other model, achieving a near-perfect accuracy of 0.9796 and an F1-score of 0.9804, high recall, and exceptional precision. Although competitive, GPT-3.5 Turbo and Alpaca Lora 7B show relatively poorer effectiveness in zero-shot and few-shot situations. The work emphasizes the effectiveness and flexibility of LLMs in a variety of linguistic circumstances, providing insightful information about the complex field of depression detection models.

\end{abstract}

\section{Introduction}

Despite being the most common mental disease in the world, depression remains an under-diagnosed disorder with potentially fatal consequences. It is an epidemic of mental illness that affects a large number of people globally. It creates a persistent sense of sorrow, a loss of interest or pleasure in activities, and can disrupt everyday tasks such as sleeping, eating, or working. Food or weight changes, difficulty sleeping, loss of energy, feelings of worthlessness or guilt, difficulty thinking, focusing, or making decisions, and thoughts of death or suicide are all symptoms of depression \citep{Mayo_Clinic_2022, WHO_2023}. Early identification and intervention are critical for effective depression management and therapy. Depression, if left untreated, can cause substantial impairments in personal, social, and professional functioning. According to the World Health Organization (WHO), the frequency of mental disease patients has increased by 13\% in the last decade. 2.8 billion of them are affected by depression, which is one of the primary causes of disability and a considerable contributor to the global sickness burden \citep{Cha_Kim_Park}.

Social media platforms are very popular as a way of self-expression in today's digital era. Individuals feel free to express their emotions and views due to the anonymity given by these platforms \citep{Maloney}. Many users often find it useful for dealing with the stresses of daily life. Platforms such as Facebook, X, and Instagram allow users to express sentiments that they would not be able to express in person. It can also have an influence on self-esteem, sleep habits, and the risk of depression, as well as anxiety symptoms. The Institute for Health Metrics and Evaluation (IHME) conducted a study from 2010 to 2019 \citep{GBD_results}, which detected depression in some age groups, revealing a significant increase in death or injury cases related to depressive disorder in Bangladesh. Approximately 4.5 million people between the ages of 20 and 54 experience depression, with the rate increasing by 1.9\% each year. Moreover, an average of 1.4 million people over the age of 50 were affected by it, with the rate increasing by 5.2\% annually.

Researchers have been experimenting with several ways to identify depressive posts on social media using machine learning and text analysis techniques, as these posts can provide insightful information about users' mental health \citep{DetectDepML,DepCEMTT}. Massive amounts of textual data can be automatically evaluated using these techniques, which have the potential to produce insightful results \citep{Aggarwal}. Specifically, NLP approaches have been used to build computer models that can identify symptoms of depression in user-generated information, like Facebook posts. These models offer a practical way to enhance current diagnostic practices and offer a cost-effective, flexible, and efficient way to screen for depression across a large population \citep{Gamon}.

Our primary objective is to precisely recognize depressive symptoms in social media profiles so that the condition can be identified early on and also determine which of the three types of language models—deep learning, conventional transformer models, and large language models—is best, given these consequences. The goal is to predict the presence of depression in Bangla social media posts with more accuracy, reduced error, and less training time. This kind of analysis has never been done on an enriched and distinct Bangla dataset because the dataset is new. This project uses large language models and advanced deep learning techniques to analyze people's social media posts to identify depression in them. Our study aims to classify individuals into two groups: those with depression and those without it, using standardized depression ratings as its basis. Several state-of-the-art (SOTA) deep learning techniques, including LSTM \citep{hochreiter1997long}, BiLSTM \citep{graves2005framewise}, GRU \citep{cho2014learning}, BiGRU \citep{chung2014empirical}, conventional transformer models like BERT Multilingual Base Model \citep{devlin2018bert}, BanglaBERT \citep{bhattacharjee2021banglabert}, sahajBERT \citep{neuropark/sahajBERT} and Bangla BERT Base \citep{sarker2020banglabert}, and large language models like GPT 3.5, GPT 3.5 turbo fine-tuned, GPT 4, and Alpaca Lora 7B have been used \citep{Ilham_Maharani_2022, Hasan_2023}.

The serious implications of untreated depression, which include mortality, underscore the importance of early care. The emphasis is on extroverted social media users in particular, as it is acknowledged that many people who struggle with depression might not publicly share their despair online. While acknowledging the limitations of using social media data, the study attempts to use the information that is now accessible to identify depressive symptoms to notify the person's immediate social circle of their emotional status.

\section{Literature Review}
\label{gen_inst}

In this section, we delve into the existing literature pertinent to our research focus. Our survey encompassed works dedicated to detecting depressive text in Bangla and various other languages. We meticulously examined and categorized these works into three primary classifications: conventional machine learning approaches, deep learning approaches, hybrid models incorporating transformers, and the integration of explainable AI with transformers, alongside extensive language-based approaches. Each of these categories is comprehensively discussed in the following sections.

\subsection{Conventional Machine Learning Approaches}

Studies using a variety of approaches have made major contributions to the field of study on depression detection from Bengali text in social media.

\citet{bhowmik2021bangla} achieved an impressive 82.21\% accuracy with Support Vector Machines (SVM) and BTSC algorithm, a novel rule-based approach for scoring text elements. Parts of speech tagging and special characters are incorporated into the BTSC algorithm, a revolutionary rule-based method for rating blogs, phrases, and words in Bengali text

\citet{shah2020hybridized} investigated several classifiers, including Naive Bayes, SVM, K-Nearest Neighbor, and Random Forest, and integrated genetic and linear approaches for feature extraction; Naive Bayes achieved an accuracy of 73.6\%. Many factors were used, including N-gram (Uni-gram, Bi-gram, and Tri-gram), TF-IDF, and linguistic features from LI-WC 2015.

\citet{hasan2018machine} applied a hybrid approach with Naive Bayes and SVM for analyzing political sentiment and achieved good results with uni-gram data.

\citet{gautam2014sentiment} used machine learning classifiers and WordNet for synonym extraction Naive Bayes achieved 88.2\%, Maximum Entropy reached 83.8\%, and Support Vector Machine achieved 85.5\% accuracy, according to the study's excellent classifier performance measures.

\subsection{Deep Learning Approaches}

Deep learning methods are better at recognizing important features and understanding the semantic context of textual input than traditional machine learning methods. Extensive research efforts have utilized state-of-the-art methods, including LSTM-CNN, RNN, hybrid models using transformers, and explainable AI combined with transformer models. 

\citet{basri2021deep} explored two model variations the first involves an attention layer fed from the output of the pooling layer through an RNN layer before reaching the output layer. In contrast, the second, named Plain CNN, excludes LSTM and attention layers, directly forwarding CNN layer outputs to the output layer. The paper incorporates a convolution layer using the soft-max function as the activation function for convolution operations. The combined CNN and RNN aim to efficiently capture and predict sentiments in Banglish texts. The models exhibit notable performance, particularly in binary classification, where CNN with LSTM and CNN with concatenated Attention and LSTM outperform multi-class classification. While the model performs well on sad and help-labeled data, challenges arise in categories like help or sarcasm due to ambiguous expressions. Research shows surprising success in identifying humor with machine learning models trained on multi-class Bengali datasets. This overcomes previous limitations in detecting humor in text.

\citet{uddin2019depression} compared GRU and LSTM recurrent neural networks for depression analysis on Bangla social media data. The study found GRU models performed better than LSTMs on this limited dataset. This offers valuable insights into the relative effectiveness of these models for depression analysis in Bengali. 

\citet{mumu2021depressed} used a hybrid CNN-LSTM model for depression detection in Bangla social media statuses. Their results varied, with accuracy ranging from 60–80\% for SVM, DT, and KNN classifiers on Facebook data and 82.2\% for an SVM model on a different dataset. GRU and LSTM models achieved better results, with 75.7\% and 88.6\% accuracy, respectively. 

\citet{shah2020early} evaluated model performance for early depression detection using metrics like Early Risk Detection Error (ERDE), Latency, and Latency-weighted F1. They found the "Word2VecEmbed+Meta" feature set performed best, achieving an F1 score of 0.81, precision of 0.78, and recall of 0.86. This highlights its effectiveness in identifying depression early.

\subsection{Hybrid and Transformer Models with Explainable AI}

\citet{ahmed2020attention} introduced an attention-based model for emotion detection in tweets. Two distinct types of embeddings, including Word2Vec, GloVe, FastText, and Sentiment Specific Word Embedding (SSWE), are employed in different components and later concatenated. The proposed model achieves a notable 79\% accuracy in emotion detection, evaluated through K-fold cross-validation with 5 folds. 

\citet{haque2020transformer} explored the use of pre-trained language models like BERT, ALBERT, ROBERTa, and XLNET for detecting suicidal ideation in social media posts. The advantage of Transformer models, particularly ROBERTa, is highlighted for surpassing conventional deep learning architectures such as BiLSTM. 

\citet{abdelwahab2022justifying} employed an attention-based LSTM (Long Short-Term Memory) model for sentiment analysis on Arabic text data, surpassing other deep learning models in accuracy. The study utilizes the Explainable AI (XAI) method, specifically LIME, to provide explanations for the sentiment classifications made by the LSTM model. The research focuses on examining posts in the Arabic text data collection that have favorable, negative, and neutral labels in order to understand consumers' overall perspectives about LASIK operations on social media.




\subsection{Large Language Models}

The landscape of mental health diagnosis is undergoing a radical transformation, thanks to the emergence of Large Language Models (LLMs) like GPT-3, Alpaca, and FLAN-T5. These AI-powered tools can analyze vast amounts of online text, potentially unlocking new avenues for identifying and treating depression. This review delves into the latest research on LLMs in this context, focusing on a comprehensive study by \cite{xu2023mental}. Their work meticulously compared diverse LLMs like Alpaca-LoRA and GPT-4 across various tasks, demonstrating that fine-tuned models like Mental-Alpaca and Mental-FLAN-T5 outperform their general counterparts by a significant margin. 

These findings showcase the immense potential of LLMs for early depression detection, but further research is crucial to address challenges like data bias and ethical considerations before full-fledged integration into clinical practice. The future promises exciting possibilities for AI-powered mental health interventions, but careful navigation is required to ensure responsible and effective implementation.

In zero- and few-shot prompting for Bangla sentiment analysis, \citet{arid2023zero} examined the efficacy of linguistic models (LLMs) such as Flan-T5, GPT-4, and Bloomz. Traditional models, such as SVM and Random Forest, function as reference points for comparison, whereas extensive pre-trained transformer models like Sentiment Analysis fine-tuning are applied to BanglaBERT, mBERT, XLM-RoBERTa, and Bloomz. BanglaBERT in particular, a monolingual transformer-based approach, regularly shows better results in sentiment analysis, surpassing other algorithms in few-shot and zero-shot situations.

\cite{kabir2023benllmeval} conducted a detailed analysis of large language models (LLMs) to see how well they performed on various Bengali natural language processing (NLP) tasks. The tasks include sentiment analysis, natural language inference, text categorization, question answering, abstractive summarization, and paraphrasing. LLMs such as Claude-2, Chat-GPT (GPT-3.5), and LLaMA-2-13b-Chat are assessed in a fine-tuning-free zero-shot learning environment. The study carefully crafts prompts for every NLP task and evaluates LLMs against cutting-edge supervised models on benchmark datasets particular to each task. Notably, ChatGPT is very good at abstractive summarization, but Claude-2 is good at answering questions.

\cite{fu2023enhancing} proposed the LLM-Counselors Support System, which supports non-professionals in providing online psychological therapies by utilizing large language models. An iterative method is used by the system to assess and potentially enhance counselor replies using AI models in order to improve communication with those who are depressed. While avoiding suggestions that were damaging or useless, the AI model concentrated on providing users with personalized guidance and help within the parameters of its assigned function. 

BERT, BART-MNLI, GPT-3 Ada, and GPT-3 Davinci are the four large language models (LLMs) based on transformer architectures that \citet{chae2023large} have thoroughly analyzed. Relative to conventional machine-learning techniques, the results show that LLMs are more accurate in recognizing stances in social media texts.

\citet{chen2023llm} investigated ChatGPT's application for chatbot simulations between psychiatrists and patients. The discoveries demonstrate that chatbots that use prompts with empathy components perform better in engagement measures and higher in empathy metrics. The chatbot with the least amount of speech shows a need for a more thorough diagnosis because of their weaker symptom recall but greater professional skills. Longer patient responses from the chatbot yield the best symptom recall, suggesting effectiveness in a comprehensive investigation of symptoms.

\citet{ji2023rethinking} investigated, with theoretical and empirical backing, the unpredictable nature of generative models in mental health prediction. The work cites a case study by \citet{yang2023mentalllama}, which underlines difficulties in comprehending complicated mental states from self-reported posts and emphasizes the value of interpretable methodologies versus potentially misleading LLM-generated explanations.

\section{Problem Description and Our Study}

Categorizing the social media published content in Bengali between two categories—depressive and non-depressive—is the root of the problem at hand. The goal is to develop a system that can conclude if a given text $\displaystyle \evx_i$ taken from a collection of $\displaystyle n$ Bengali social media texts $\displaystyle \rmX$ = {\{$\displaystyle \ervx_1$, $\displaystyle \ervx_2$, $\displaystyle \ervx_3$,.....$\displaystyle \ervx_n$}\} belongs to one of the two predefined categories: $\displaystyle \rmC$ = {\{$\displaystyle \ervc_1$, $\displaystyle \ervc_2$}\}. The system's main goal is to classify each text as $\displaystyle \mC_i$, where depressed and non-depressive texts are indicated by the labels $\displaystyle \ervc_1$ and $\displaystyle \ervc_2$, respectively.

In this research effort, we created the Bengali Social Media Depressive Dataset (BSMDD), a comprehensive dataset of depressive content sourced from Reddit and X (formerly Twitter). Strenuous efforts were taken to ensure the cleanliness of the dataset by eliminating duplicates and maintaining high translation and annotation quality standards. This dataset is now publicly accessible to the wider research community.

Our approach to experimentation goes beyond the conventional approaches of deep learning and transformer models that are commonly used. Specifically, we delved into the efficacy of sophisticated models, including GPT-3.5 Turbo, our customized model DepGPT (fine-tuned on GPT-3.5), GPT-4, and Alpaca Lora 7B, exploring their performance in both zero-shot and few-shot settings. The results of this experiment contribute valuable insights to understanding the detection of depressing content in social media.

\section{Dataset}

\subsection{Dataset Collection}


Our experiment is based on two well-established datasets that are commonly employed for mental health analysis collected from social media platforms. A total of 31,695 English textual samples were amassed from reputable sources, namely Reddit \citep{Reddit} and X (formerly Twitter)  \citep{Sentiment140} verified datasets, which native speakers had meticulously scrutinized (Table \ref{tab:Dataset-table}). Notably, we intentionally avoid using texts that contain less than 30 words as it would be hard to understand emotions in these texts.

\textbf{Reddit dataset \citep{Reddit}}: We utilized a dataset sourced from the Reddit community, specifically focusing on submissions from two relevant subreddits: r/depression and r/SuicideWatch. The dataset comprises a total of 20,000 submissions. We employed this dataset for the depressive texts in our datasets.

\textbf{Sentiment140 dataset \citep{Sentiment140}}: For the non-depressive component of our dataset, we leveraged the Sentiment140 dataset, a comprehensive collection of 1,600,000 tweets extracted using the Twitter API. We have collected approximately 11,000 datasets from this.

\subsection{Annotations \& Translations}

The datasets we gathered from Reddit \citep{Reddit} and X (formerly Twitter) \citep{Sentiment140} were categorized into depressive and non-depressive segments.

To ensure the precision of translations from English to Bengali, seven (7) persons participated in the translation and annotation process. These persons are native Bengali speakers and are proficient in both Bengali and English. They are also currently pursuing studies in both Bengali language and computer science. In addition to their academic qualifications, they have also gained experience in the domain of mental health by participating in several psychological groups on social media. Participating in this group has helped them develop a deeper understanding of the different emotions and experiences associated with mental health. This knowledge has been valuable in their work on the translation and annotation project, as it has helped them to accurately identify and label the emotions expressed in the text.

The final translation assigned to each post was determined through a collaborative approach, with consensus among annotators being the determining factor. In instances of disagreement, a consensus meeting was convened to address disparities and reach a conclusive decision. In this way, we compiled a comprehensive dataset consisting of content related to depression, known as the Bengali Social Media Depressive Dataset (BSMDD).

It's worth noting that participants in the annotation process were remunerated for their contributions, acknowledging the effort and expertise invested in ensuring the quality of the annotated dataset.

\begin{table}[h!]
\centering
\caption{Dataset characterization table}
\label{tab:Dataset-table}
\begin{tabular}{c|c}
\hline
{\color[HTML]{000000} Language type} & Bangla          \\
Platforms                            & Reddit, Kaggle  \\
Gender biases                        & Male and female \\
Total amassed samples                & 31,695          \\
Total Words                          & 59,23,168       \\
Average no. of words in a sample     & 187             \\
Total characters                     & 3,38,13,794     \\
Average characters in a sample       & 1069            \\ \hline
\end{tabular}
\end{table}

\section{Proposed Methodology}


\begin{figure}[h!]
    \centering
    \includegraphics[width=0.8\textwidth]{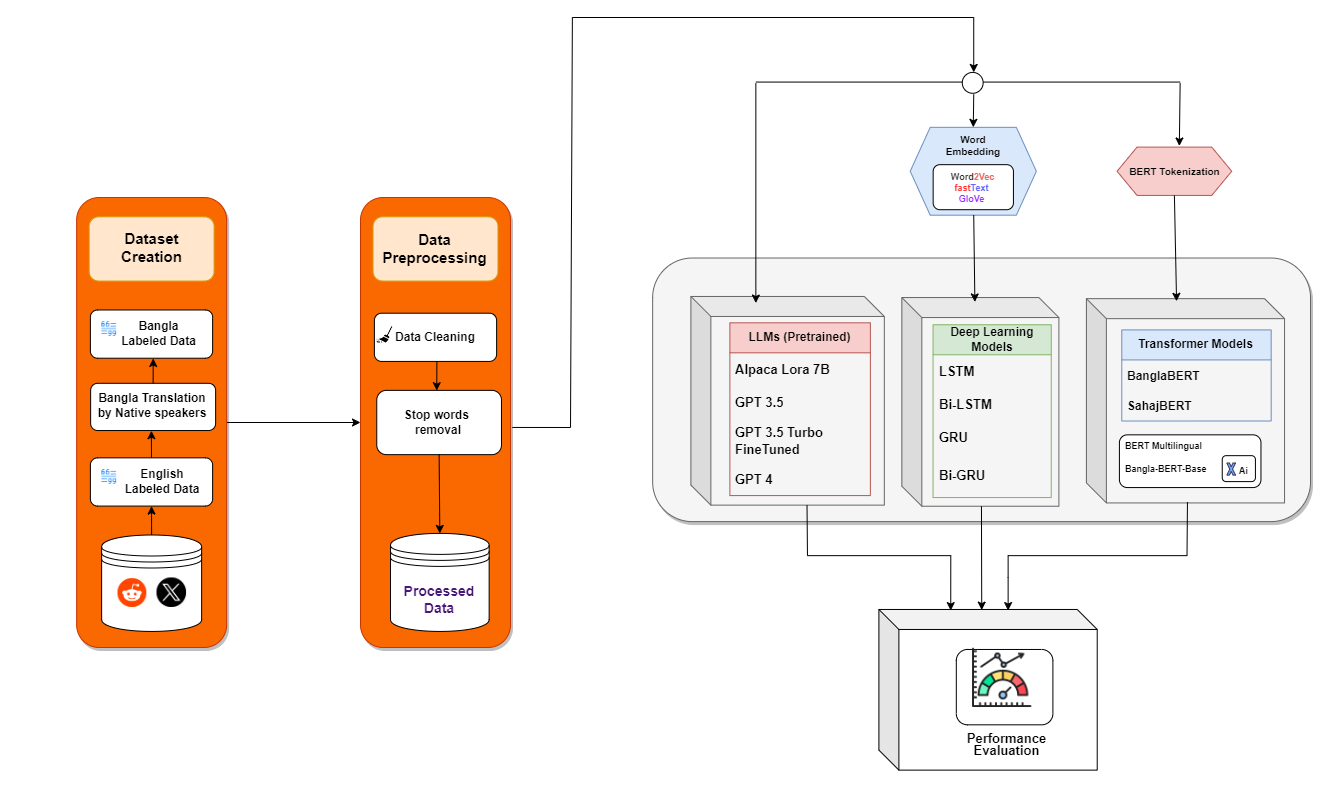} 
    \caption{Proposed Methodology} 
    \label{fig:Methodology}
\end{figure}

\subsection{Data Preprocessing}

The texts obtained after translation from Reddit and Twitter frequently contain elements that introduce noise, such as usernames, hashtags, URLs, English characters, and symbols. To enhance the data quality, we applied several preprocessing filters to the collected texts. As a result of these filters, we were able to remove 3,695 texts, leaving 28,000 texts that were then sent to embedding and experimental evaluations. The following steps were taken to process the texts:
\begin{itemize}
\item Duplicate texts were removed.
\item Repeated punctuations were removed.
\item Short texts (less than 30 words) were excluded as it's difficult to grasp emotions from such brief texts.
\item English characters and numbers were removed using nltk\citep{NLTK}.
\item Bengali text was segmented into tokens and stop words were removed using BNLTK \citep{bnltk}.
\item The Bengali texts were reduced to their root forms using bangla-stemmer \citep{bnst}.
\end{itemize}

We divided our dataset of 28,000 texts into two categories: 14,000 “Depressive” and 14,000 “Non-Depressive” for the evaluation of deep learning and large-scale pre-trained transformer models (PLMs) (see Table \ref{tab:Dataset-Preprocessing}).

\begin{table}[h!]
\centering
\caption{Dataset Summary After Preprocessing}
\label{tab:Dataset-Preprocessing}
\begin{tabular}{l|l}
\hline
Total processed samples    & 28,000 \\
Depressive samples (1)     & 14,000 \\
Non-Depressive samples (0) & 14,000 \\
Training samples           & 22,400 \\
Testing samples            & 5,600  \\ \hline
\end{tabular}
\end{table}

\begin{table}[h!]
\centering
\caption{Dataset summary for Large Language Model}
\label{tab:Dataset-LLM-table}
\begin{tabular}{cccc}
\hline
\textbf{Class} & \textbf{Train} & \textbf{Test} & \textbf{Text Length} \\ \hline
Depressive     & 43.837K        & 19.6K         & 511+-74              \\
Non Depressive & 43.836K        & 16.1K         & 525+-60              \\
Total          & 87.675K        & 35.7K         &                      \\ \hline
\end{tabular}
\end{table}

For assessing large language models, we further categorized the data into training and test subsets based on tokens. The "Depressive" class had approximately 43,837K training tokens and 19.6K testing tokens, with an average token length of 511 characters. The "Non Depressive" class had around 43,836K training tokens and 16.1K testing tokens, with an average token length of 525 characters. In total, the dataset contained 87.675K tokens, distributed across training and test sets for both classes. This division provides insights into token sample distribution and average lengths for further analysis, as illustrated in Table \ref{tab:Dataset-LLM-table}

\subsection{Models}

We conducted our experiments using deep learning models, large-scale pre-trained transformer models (PLMs), and large language models (LLMs). For the performance measure for all different experimental settings, we compute accuracy, precision, recall, and F1 score. We split the dataset into 80:20 and got 22,400 texts for training and 5,600 texts for training (show in Table \ref{tab:Dataset-Preprocessing}).

\subsubsection{Deep Learning Models}

While deep learning models such as LSTM \citep{hochreiter1997long}, BiLSTM \citep{graves2005framewise}, GRU \citep{cho2014learning}, BiGRU \citep{chung2014empirical} have been widely used in prior studies and remain in use of sentiment analysis, we also wanted to assess their performance on our dataset. We trained the models using the training set, fine-tuned the parameters with the development set, and assessed the model’s performance on the test set To prepare the data for these models, we transformed the text into a word embedding representation, using Word2vec, GloVe, and fastText. We present the architecture details of all the deep learning models in Table \ref{tab:DeepLearning-Architecture}.

\begin{enumerate}
    \item[(a)] LSTM: Long Short-Term Memory is a foundational deep learning architecture designed to overcome the vanishing gradient problem in recurrent neural networks. LSTMs maintain a cell state that can selectively store or remove information, allowing them to capture long-range dependencies in sequential data. This makes them well-suited for tasks such as language modelling and sentiment analysis, where understanding the context of words across extended sequences is crucial. LSTMs consist of memory cells and gates that regulate the flow of information, enabling effective learning of temporal patterns. Here are the main equations along with explanations of the variables:

        \begin{align}
            i_t &= \sigma(W_{ii} \cdot x_t + b_{ii} + W_{hi} \cdot h_{t-1} + b_{hi}) \label{eq:input_gate} \\
            f_t &= \sigma(W_{if} \cdot x_t + b_{if} + W_{hf} \cdot h_{t-1} + b_{hf}) \label{eq:forget_gate} \\
            o_t &= \sigma(W_{io} \cdot x_t + b_{io} + W_{ho} \cdot h_{t-1} + b_{ho}) \label{eq:output_gate}
        \end{align}      
        
    In these equations:  
    \begin{align*}
        & x_t \text{ is the input at time step } t, \\ 
        & f_t \text{ represents the forget gate output at time step } t, \\
        & h_{t-1} \text{ is the hidden state from the previous time step,} \\
        & W \text{ and } b \text{ are the weights and biases,} \\
        & \sigma \text{ is the sigmoid activation function} \\
    \end{align*}

    \item[(b)] BiLSTM: Bidirectional Long Short-Term Memory extends the LSTM architecture by processing input sequences in both forward and backward directions simultaneously. This bidirectional approach enhances the model's capability to understand the context by considering information from past and future time steps. BiLSTMs, like LSTMs, leverage memory cells and gates, allowing them to capture and update information across different time steps bidirectionally. They are particularly effective for text classification tasks where word meanings depend on the surrounding words in both directions, providing a comprehensive understanding of sequential data. Here are the forward and backward equations, along with explanations of the variables:

    \item[(c)] GRU: Gated Recurrent Unit stands as a robust architecture in the realm of sentiment analysis. GRUs, an enhancement of traditional recurrent neural networks, are adept at capturing dependencies in sequential data. Unlike standard RNNs, GRUs introduce gating mechanisms that enable the model to selectively update its memory, facilitating better long-range dependency handling. In the context of sentiment analysis, GRUs prove valuable for capturing nuanced contextual information and understanding the emotional flow within a given text. The model's ability to selectively retain relevant information ensures efficient sentiment representation. Equations are as follows:
    \begin{align}
        z_t &= \sigma(W_{z} \cdot x_t + U_{z} \cdot h_{t-1} + b_{z}) \\
        r_t &= \sigma(W_{r} \cdot x_t + U_{r} \cdot h_{t-1} + b_{r}) \\
      \tilde{h}_t &= \tanh(W_{h} \cdot x_t + U_{h} \cdot (r_t \odot h_{t-1}) + b_{h}) \\
        h_t &= (1 - z_t) \odot h_{t-1} + z_t \odot \tilde{h}_t 
    \end{align}
    
    In these equations:
    \begin{align*}
        & \tanh \text{ is the hyperbolic tangent activation function.}
    \end{align*}
    
    \item[(d)] BiGRU: Building upon the foundation of GRUs, a Bidirectional Gated Recurrent Unit emerges as a formidable architecture for sentiment analysis tasks. BiGRUs process input sequences in both forward and backward directions concurrently, allowing the model to assimilate information from past and future contexts. This bidirectional approach enhances the model's comprehension of sentiment nuances by considering the entire context surrounding each word. The memory cells and gating mechanisms in BiGRUs contribute to a more comprehensive understanding of sentiment, making them particularly effective for discerning subtle emotional cues in text data. Here are the forward and backward equations, along with explanations of the variables:

\end{enumerate}

\begin{table}[h!]
\centering
\caption{Information on Deep Learning Models' Architecture}
\label{tab:DeepLearning-Architecture}
\begin{tabular}{c|cccc}
\hline
\textbf{Architecture}     & \textbf{LSTM} & \textbf{BiLSTM} & \textbf{GRU} & \textbf{BiGRU} \\ \hline
Input length              & \multicolumn{4}{c}{512}                                         \\ \hline
Embedding Dimension       & \multicolumn{4}{c}{300}                                         \\ \hline
Bidirectional             & No            & Yes             & No           & Yes            \\
Dropout                   & -             & 0.5             & -            & 0.5            \\
Total Parameters          & 10,288,202    & 7,799,998       & 8,777,751    & 8,049,198      \\ \hline
Activation Function       & \multicolumn{4}{c}{ReLU}                                        \\ \hline
Activation (Output Layer) & \multicolumn{4}{c}{Softmax}                                     \\ \hline
Loss Function             & \multicolumn{4}{c}{Categorical Crossentropy}                    \\ \hline
Optimizer                 & \multicolumn{4}{c}{Adam}                                        \\ \hline
\end{tabular}
\end{table}

\newpage

\subsubsection{Large-scale Pre-trained Transformer Models (PLMs)}

Large-scale pre-trained transformer models (PLMs) have achieved state-of-the-art performance across numerous NLP tasks. In our study, we fine-tuned several of these models. These included the BERT Multilingual Base Model \citep{devlin2018bert}, the monolingual transformer model BanglaBERT \citep{bhattacharjee2021banglabert}, sahajBERT \citep{neuropark/sahajBERT} and Bangla BERT Base \citep{sarker2020banglabert}. We fine-tuned each model using the default settings over 10 to 50 epochs and 8 to 128 batch sizes.

\begin{enumerate}
    \item[(a)] BERT Base: A deep learning framework called Bidirectional Encoder Representations from Transformers (BERT) links input and output elements and adaptively assigns weights based on their relationship. One of BERT's unique selling points is its capacity for bidirectional training, which allows the language model to understand a word's context by taking into account words that are close to it instead of just concentrating on the word that comes before or after it. BanglaBERT Base \citep{sarker2020banglabert} adheres to the same architecture as the original BERT model. The techniques used in BERT are as follows:

    \subsubsubsection{Self-Attention Mechanism:}
    \begin{equation}
        \text{Attention}(Q, K, V) = \text{softmax}\left(\frac{QK^T}{\sqrt{d_k}}\right) V
    \end{equation}
    \subsubsubsection{Multi-Head Self-Attention:}
    \begin{align}
        \text{MultiHead}(X) = \text{Concat}(\text{head}_1, \ldots, \text{head}_h) W_O \\
        \text{head}_i = \text{Attention}(XW_{Qi}, XW_{Ki}, XW_{Vi})
    \end{align}
    \subsubsubsection{Layer Normalization and Feedforward Network:}
    \begin{align}
        \text{Output} = \text{FeedForward}(\text{LayerNorm}(\text{MultiHead}(X))) \\
        \text{FeedForward}(X) = \text{ReLU}(X W_{FF1} + b_{FF1}) W_{FF2} + b_{FF2}
    \end{align}
    Here,
    \begin{align*}
        & X \text{ is the input sequence of embeddings,} \\
        & Q, K, V \text{ are query, key, and value matrices,} \\
        & d_k \text{ is the dimensionality of key vectors,} \\
        & W \text{ and } b \text{ are the weights and biases,} \\
    \end{align*}
    
    \item[(b)] ELECTRA based: Efficiently Learning an Encoder that Classifies Token Replacements Accurately (ELECTRA) \citep{clark2020electra} identifies replaced tokens in the input sequence. To do this, a generator model must be trained to predict the original tokens for the masked-out ones, and a discriminator model must be trained to distinguish between original and substituted tokens. ELECTRA discriminator model is BanglaBERT \citep{bhattacharjee2021banglabert}.
    
    \item[(c)] ALBERT based: Larger models are not always necessary to achieve improved language models, as shown by ALBERT (A Lite BERT) \citep{lan2019albert}. Three major changes are made to the original Transformer's encoder segment architecture in order to do this: factorized embedding parameters, cross-layer parameter sharing, and the use of Sentence-order prediction (SOP) rather than Next Sentence Prediction (NSP). Using masked language modeling (MLM) and Sentence Order Prediction (SOP) aims, sahajBERT \citep{neuropark/sahajBERT} is a cooperatively pre-trained ALBERT model in the Bengali language context.

\end{enumerate}

\begin{table}[h!]
\centering
\caption{Information on Transformer Models' Architecture}
\label{tab:Transformer-Architecture}
\begin{tabular}{ccccc}
\hline
\textbf{Architecture} & \textbf{Used Model} & \textbf{Layer} & \textbf{Hyperparameter} & \textbf{Parameter} \\ \hline
BERT & BERT Base       & 12 & 12 & 110M \\
                      & BanglaBERT Base & 12 & 12 & 110M \\
ELECTRA               & BanglaBERT      & 12 & 12 & 110M \\
ALBERT                & sahajBERT       & 24 & 16 & 18M  \\ \hline
\end{tabular}
\end{table}

\subsubsection{ Large Language Models (LLMs)}

Based on the quantity of parameters that each model was trained on, the large language models (LLMs) from Google, Meta, and OpenAI were compared in this study. Particularly when it comes to LLMs, parameters play a crucial role as they are often correlated with the model's comprehension and production of language that is human-like. A model with more parameters is often better able to learn from large volumes of data, leading to more accurate and sophisticated language creation and interpretation.

\textbf{GPT Models}
\begin{enumerate}
    \item[(a)] GPT-3.5: It is OpenAI's large language model (LLM). It is an improved version of GPT-3, which underwent extensive training on a large text and code dataset. The tasks that GPT-3.5 can complete include text creation, translation, summarization, answering questions, and creative writing\citep{Ye_2023}.
    \item[(b)] GPT-4: It is a multi-modal model, it can process both text and image input, produce text, translate between languages, create various types of creative content, and provide answers to questions. More complicated tasks can be handled by GPT-4 than by earlier GPT models. The model performs at human levels across a wide range of academic and professional benchmarks\citep{OpenAI_2023}.
    \item[(c)] DepGPT: It is a refined version of GPT-3.5 Turbo fine-tuned model which is designed for specific tasks and trained to follow commands thoughtfully. Fine-tuning is crucial for maximizing performance in big language models, allowing developers to personalize the model with domain-specific data. This tailoring enhances relevance, accuracy, and performance for niche applications, enabling the creation of personalized AI applications\citep{Latif_2023}. Fine-tuning also increases steer-ability and dependability, making the model more predictable in production. Additionally, it significantly improves performance, allowing DepGPT to potentially surpass GPT-4 capabilities in specialized tasks.
\end{enumerate}

GPT-4 and GPT-3.5 were also tested using typical machine learning benchmarks. GPT-4 beats existing big language models as well as the majority of state-of-the-art (SOTA) models, which may incorporate benchmark-specific programming or extra training methods.

\textbf{LLaMA Models}

Stanford University's Alpaca 7B model is a small, efficient instruction-following language model that mimics OpenAI's text-davinci-003 but is more cost-effective for academic research. It was fine-tuned using Text-davinci-003 and performed better than OpenAI's text-based model. However, Alpaca 7B has limitations like generating erroneous information and perpetuating societal preconceptions. Low-Rank Adaptation (LoRA) approaches, such as the Stanford Alpaca cleaned version dataset, can help overcome these limitations without sacrificing performance. \citep{andermatt2023uzh_pandas}

LoRA Low-Rank Adaptation (LoRA) was utilized in the Alpaca 7B model to enhance efficiency in natural language processing, reducing trainable parameters to 16 million while maintaining model performance and reducing processing needs \citep{hu2021lora}.


\begin{table}[h!]
\centering
\caption{Information of Large Language Models’ Architecture}
\label{tab:LLM-Architecture}
\resizebox{\columnwidth}{!}{%
\begin{tabular}{|c|cc|c|}
\hline
\textbf{Features / Models} &
  \multicolumn{1}{c|}{\textbf{Parameter size}} &
  \textbf{Structure} &
  \textbf{Architecture} \\ \hline
GPT-3.5 &
  \multicolumn{1}{c|}{175B} &
  Decoder-only, 12 layers &
  \begin{tabular}[c]{@{}c@{}}Autoregressive, masked \\ language modeling\end{tabular} \\ \hline
GPT-3.5 Finetuned &
  \multicolumn{2}{c|}{Same as GPT-3.5} &
  Fine-tuned for specific tasks \\ \hline
GPT-4 &
  \multicolumn{1}{c|}{1.7T} &
  120 layers &
  \begin{tabular}[c]{@{}c@{}}Multimodal (text and image), \\ likely autoregressive\end{tabular} \\ \hline
Alpaca 7B &
  \multicolumn{1}{c|}{7B} &
  Encoder-decoder, 6 layers each &
  \begin{tabular}[c]{@{}c@{}}Sequence to  sequence, \\ masked \& factual language \\ modeling\end{tabular} \\ \hline
\end{tabular}%
}
\end{table}

We used OpenAI’s playground to execute our tasks, which has mainly 3 sections: System, User, and Assistant. The 'System' section specifies how the model will respond. It is essentially a collection of exact instructions for specific formatting that is communicated to the model with each prompt. The 'User' section contains the main prompt input into the model, while the 'Assistant' section contains the model's response. Mode, Model, Top P, Frequency penalty, Presence penalty, and Stop sequence are the other controls.

\textbf{Prompt Design}

To provide accurate instructional prompting to perform a specific task, the model makes use of system prompts. Regarding our categorization of depression, the system prompt was instructed to function as an expert psychologist with extensive knowledge of recognizing depression in Bangla literature. It will examine every text and evaluate it word for word before categorizing it as either "Non-Depressive" or "Depressive." Direct instructions combined with user prompts are used to categorize the provided Bangla text. The prompt begins with a multiple-choice binary question, and an equal symbol indicates the content that follows. It is best to either completely remove or set to null the input portion prompting from the interface. The assistant provided the classification result in response to the model.

Three models from OpenAI were utilized: GPT-3.5 Turbo, GPT-4, and a fine-tuned version of GPT-3.5 Turbo using our dataset that we have named \textbf{DepGPT}. The models were chosen above alternative models due to their cost-effectiveness and efficient tokenization technique, which reduces the number of tokens for Bangla words. As a result, API costs are also minimized. Instead of detecting many tokens in English, the default GPT-3 tokenizer (see Figure 5.4) finds many tokens in Bangla. As a result, the token size grows with each command. The price of the API rises as a result, which, given our limited means, is neither practical nor economical.

\textbf{Zero-Shot}

\begin{table}[h!]
\centering
\caption{Design of a Zero Shot Sample}
\label{tab:Zero-Shot sample}
\begin{tabular}{|l|l|}
\hline
Instruction & System \\
\hline
Task & User prompt \\
\hline
Assistant & Output will be generated by LLM \\
\hline
\end{tabular}
\end{table}

\begin{figure}[h!]
    \centering
    \caption{Design of Zero Shot Example}
    \includegraphics[width=.8\textwidth]{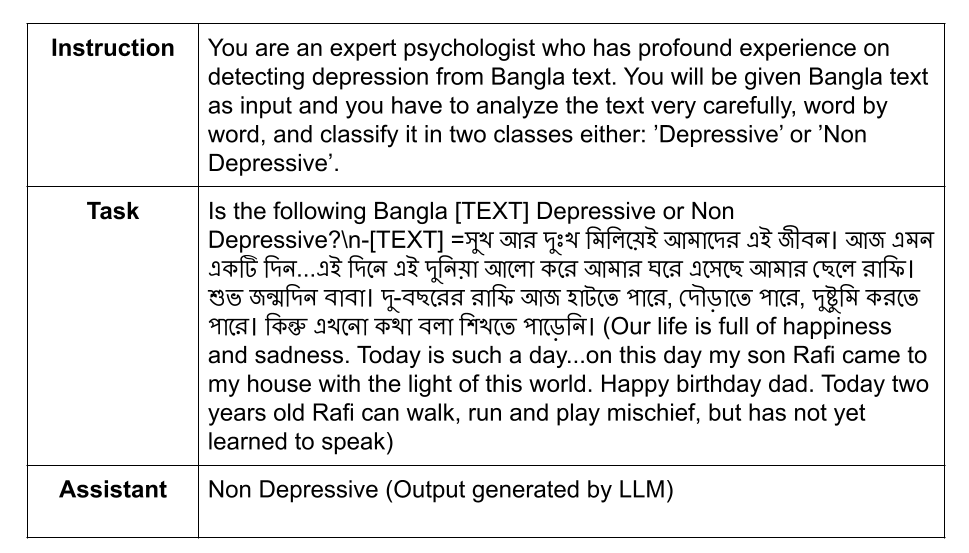} 
    \label{fig:Zero-Shot Example}
\end{figure}

We used a zero-shot learning technique that operates on classes or data that are not explicitly present in the model's training set. This method allows the model to anticipate or perform tasks associated with novel classes or tasks, demonstrating its capacity to generalize beyond specific training cases. We used the format shown in Table \ref{tab:Zero-Shot sample} to organize the zero-shot example prompt in our technique. We gave the model an "instruction" and a Task as inputs, and we expected the "Assistant" to produce an output categorizing the text as either "Depressive" or "Non-Depressive." Importantly, the model completed this job without any prior context knowledge, demonstrating its capacity to classify objects based only on the task and instruction given, thus capturing the spirit of zero-shot learning shown in Figure \ref{fig:Zero-Shot Example}.

\textbf{Few-Shot}

When compared to the zero-shot learning scenario, few-shot learning performs better, as demonstrated by the groundbreaking work of \citep{brown2020language}. Numerous benchmarking studies \citep{Ahuja} have also confirmed this. In our methodology, we have employed few-shot learning to tackle the challenge of learning from limited labelled data. Unlike traditional machine learning paradigms that often necessitate substantial labelled datasets for training, few-shot learning focuses on training models capable of generalizing and making accurate predictions even when presented with only a small number of examples per class. This approach proves particularly advantageous in scenarios where acquiring extensive labelled data is either cost-prohibitive or unfeasible. The format of our few-shot prompt structure is shown in Table \ref{tab:few-shot-example-table}.

\begin{table}[h!]
\centering
\caption{Design of a Few Shot Sample}
\label{tab:few-shot-example-table}
\begin{tabular}{|l|l|}
\hline
Instruction                                              & : \{System prompt\}                                                                            \\ \hline
\begin{tabular}[c]{@{}l@{}}Task-1\\ Assistant\end{tabular} & \begin{tabular}[c]{@{}l@{}}: \{User prompt {[}Depressive{]}\}\\ : \{Deppressive\}\end{tabular} \\ \hline
\begin{tabular}[c]{@{}l@{}}Task-2\\ Assistant\end{tabular} & \begin{tabular}[c]{@{}l@{}}: \{User prompt {[}NonDepressive{]}\}\\ : \{Non Depressive\}\end{tabular} \\ \hline
\begin{tabular}[c]{@{}l@{}}Task-3\\ Assistant\end{tabular} & \begin{tabular}[c]{@{}l@{}}: \{User prompt\}\\ : \{Output will be generated by LLM\}\end{tabular}  \\ \hline
\end{tabular}
\end{table}

\begin{figure}[h!]
    \centering
    \caption{Design of Few Shot Example}
    \includegraphics[width=.9\textwidth]{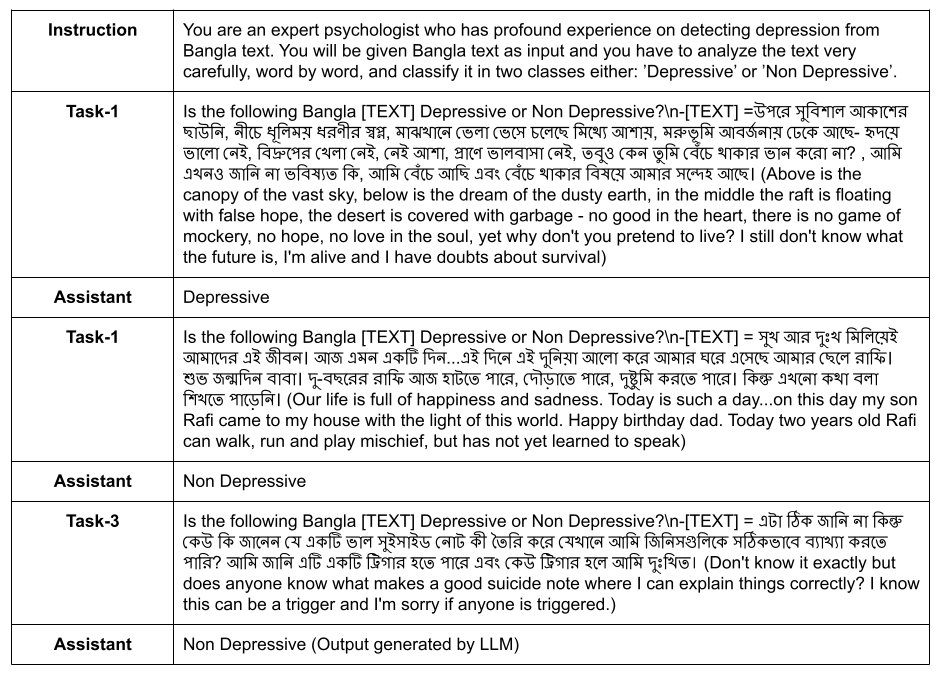} 
    \label{fig:Few-Shot Example}
\end{figure}

The example prompt was set up as depicted in Figure \ref{fig:Few-Shot Example} above. It consisted of an "Instruction", Tasks-1 and Task-2, and their corresponding responses from the model via "Assistant". The model received input from both samples, one of which was "Depressive" and the other "Non Depressive". Next, the last task was included at the conclusion of the prompt and forwarded to the model. 'Assistant' is the final solution that categorizes Task 3 using the provided examples and previous knowledge.

\textbf{Other controls}

Temperature, Top P, Maximum Tokens, Frequency penalty, Presence penalty, and stop sequence are respectively used to control the output’s randomness, diversity, generative token limits, and the amount of penalization on new tokens based on their existing frequency and appearance in the texts. The configurations used are given below:

\begin{table}[h!]
\centering
\caption{Control of Large Languge Model}
\label{tab:LLM-control-table}
\begin{tabular}{cc}
\hline
Temperature       & 1.0 \\
Top P             & 1.0 \\
Maximum tokens    & 256 \\
Frequency penalty & 0.0 \\
Presence penalty  & 0.0 \\ \hline
\end{tabular}
\end{table}

\section{Result and Discussion}

In this summary, we present the outcomes of our experiment employing various deep learning models (Section 6.1), transformer models (Section 6.2), and large language models (Section 6.3). While our primary focus revolves around tasks related to detecting depression in this study, we also implemented explainable AI to transformer models in Section 6.2.

In essence, our findings reveal promising performance in zero-shot and few-shot scenarios using large language models (LLMs) for depression detection tasks. However, this performance is still constrained. Notably, fine-tuning the models with instructions on the dataset significantly enhances their performance across all tasks concurrently. Our detailed examination highlights the robust reasoning capabilities of specific LLMs, particularly GPT-3.5 and GPT-4. It is crucial to note that these results do not imply deployability. In Section 6, we underscore important ethical considerations and identify limitations in our research.

\subsection{Evaluation of Deep Learning Models}

GRU, BiGRU, LSTM, and BiLSTM are fundamental deep-learning models used in sentiment analysis due to their ability to handle sequential data and capture long-term dependencies. These models use gating mechanisms to update their hidden state, retaining past information relevant to the current time step. BiGRU and BiLSTM can capture data from both forward and backward time series and handle the vanishing gradient problem well.  This study systematically assesses the performance of three-word embeddings Word2Vec, Glove, FastText and four deep learning architectures LSTM, Bi-LSTM, GRU, and Bi-GRU in depression detection tasks.

Deep learning models were used to predict text depressive and non-depressive, with data split into 80:20 sets for training and testing, and experimented with different layer counts(0–3), drop out(0.2-0.5), weight decay(0.001 to 0.01) and batch size (8,16,32,64).
The accuracy results are summarised in Table \ref{tab:result-Deep-Learning}.

\begin{table}[h!]
\centering
\caption{Performance of the Deep Learning Architectures}
\label{tab:result-Deep-Learning}
\resizebox{\columnwidth}{!}{%
\begin{tabular}{c|cccccc|cc}
\hline
\textbf{\begin{tabular}[c]{@{}c@{}}Models \& \\ Embeddings\end{tabular}} &
  \textbf{\begin{tabular}[c]{@{}c@{}}Batch \\ Size\end{tabular}} &
  \textbf{Epoch} &
  \textbf{\begin{tabular}[c]{@{}c@{}}Learning \\ Rate\end{tabular}} &
  \textbf{Dropout} &
  \textbf{\begin{tabular}[c]{@{}c@{}}Weight \\ Decay\end{tabular}} &
  \textbf{Layers} &
  \textbf{\begin{tabular}[c]{@{}c@{}}Accuracy \\ (Maximum)\end{tabular}} &
  \textbf{\begin{tabular}[c]{@{}c@{}}ROC-AUC \\ score\end{tabular}} \\ \hline
LSTM (Word2Vec)    & 32 & 20 & 0.001 & 0.2 & 0.001 & 3 & 0.8722          & 0.851 \\
LSTM (fastText)    & 32 & 25 & 0.01  & 0.2 & 0.001 & 4 & 0.887           & 0.879 \\
LSTM (GloVe)       & 16 & 10 & 0.001 & 0.2 & 0.001 & 2 & 0.88            & 0.879 \\
Bi LSTM (Word2Vec) & 32 & 20 & 0.001 & 0.2 & 0.005 & 3 & 0.885           & 0.874 \\
Bi LSTM (fastText) & 64 & 20 & 0.001 & 0.2 & 0.001 & 3 & 0.8987          & 0.860 \\
Bi LSTM (GloVe)    & 16 & 10 & 0.001 & 0.3 & 0.001 & 2 & 0.882           & 0.880 \\
GRU (Word2Vec)     & 16 & 15 & 0.001 & 0.3 & 0.001 & 2 & 0.8948          & 0.881 \\
GRU (fastText)     & 64 & 25 & 0.001 & 0.4 & 0.01  & 2 & 0.9002          & 0.882 \\
GRU (GloVe)        & 16 & 20 & 0.001 & 0.2 & 0.001 & 3 & 0.8777          & 0.884 \\
Bi GRU (Word2Vec)  & 16 & 15 & 0.001 & 0.3 & 0.001 & 3 & \textbf{0.8627} & 0.853 \\
Bi GRU (fastText)  & 16 & 15 & 0.001 & 0.3 & 0.005 & 4 & \textbf{0.9036} & 0.901 \\
Bi GRU (GloVe)     & 16 & 15 & 0.001 & 0.2 & 0.001 & 2 & 0.8791          & 0.853 \\ \hline
\end{tabular}%
}
\end{table}

As depicted in Table \ref{tab:result-Deep-Learning}, the models are evaluated under varying hyperparameters. Key findings include:

Bi-GRU with fastText embeddings achieved the highest accuracy of 90.36\%, outperforming other models, including LSTM and Bi-directional LSTM, across Word2Vec, fastText, and GloVe embeddings.  The full table of \ref{tab:result-Deep-Learning} is shown in \ref{tab:result-Deep-Learning2}.

\subsection{Evaluation of Large-scale Pre-trained Transformer Models (PLMs)
}

In this series of experimental setups, Large-scale pre-trained transformer models (PLMs) like the BERT Multilingual Base Model, monolingual transformer model BanglaBERT, sahajBERT, and BERT-Base-Bangla were assessed, each involving a distinct set of critical hyperparameters. These hyperparameters included batch size (8-128), learning rate (0.01 - 0.0005), number of epochs (10-50), number of folds (5), momentum (0.9), and dropout (0.1), all of which had significant implications for model performance.
The metrics include accuracy, precision, recall, and F1 score. Among the models, sahajBERT demonstrates the highest performance across all metrics, with an accuracy of 0.867, precision of 0.8718, and F1 score of 0.8662.

The maximum performance of the large-scale pre-trained transformer models is illustrated in the table \ref{tab:result-transformer}. The full table of \ref{tab:result-transformer} is shown in \ref{tab:result-bert-multi-table2},\ref{tab:result-BanglaBERT-table2},
\ref{tab:result-sahajBERT-table2},
\ref{tab:result-Bangla-BERT-Base-table2}.

\begin{table}[h!]
\centering
\caption{Performance of the Large-scale Pre-trained Transformer Models}
\label{tab:result-transformer}
\begin{tabular}{|c|c|c|c|c|c|c|c|}
\hline
\textbf{Models} &
  \textbf{\begin{tabular}[c]{@{}c@{}}Batch \\ Size\end{tabular}} &
  \textbf{\begin{tabular}[c]{@{}c@{}}Learning \\ Rate\end{tabular}} &
  \textbf{\begin{tabular}[c]{@{}c@{}}Num of \\ epoch\end{tabular}} &
  \textbf{\begin{tabular}[c]{@{}c@{}}Accuracy\\ (Max)\end{tabular}} &
  \textbf{Precision} &
  \textbf{Recall} &
  \textbf{F1 Score} \\ \hline
\begin{tabular}[c]{@{}c@{}}BERT Multilingual \\ Base Model (Cased)\end{tabular} & 32  & \multirow{4}{*}{0.01} & 10 & 0.8233         & 0.7935 & 0.8986 & 0.8544 \\ \cline{1-2} \cline{4-8} 
Bangla-Bert-Base                                                                & 128 &                       & 40 & 0.8535         & 0.8405 & 0.8967 & 0.8582 \\ \cline{1-2} \cline{4-8} 
BanglaBERT                                                                      & 64  &                       & 20 & 0.8604         & 0.8692 & 0.8932 & 0.8625 \\ \cline{1-2} \cline{4-8} 
sahajBERT                                                                       & 128 &                       & 13 & \textbf{0.867} & 0.8718 & 0.8816 & 0.8662 \\ \hline
\end{tabular}
\end{table}

\subsubsection{Explainability in the feature level}

\begin{figure}[h!]
    \centering
    \includegraphics[width=0.9\textwidth]{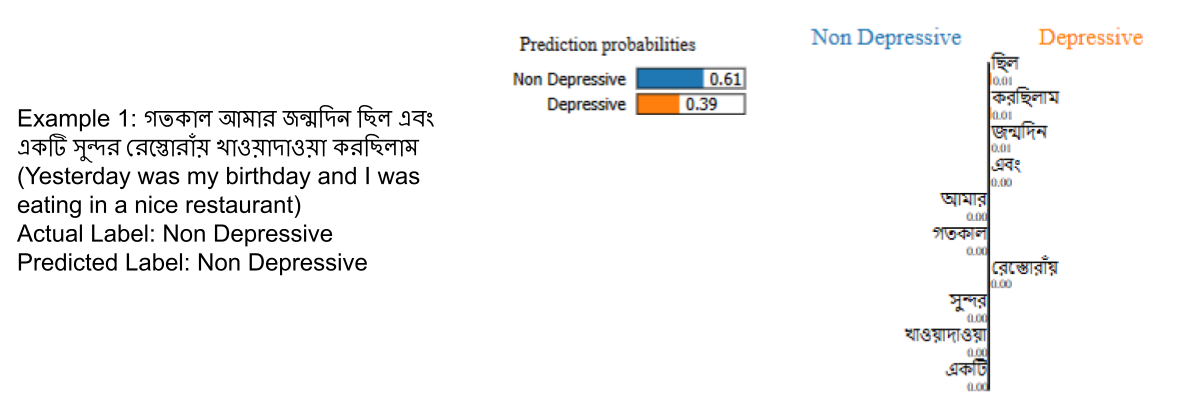} 
    \caption{A true positive sample example} 
    \label{fig:True_Positive}
\end{figure}

\begin{figure}[h!]
    \centering
    \includegraphics[width=0.9\textwidth]{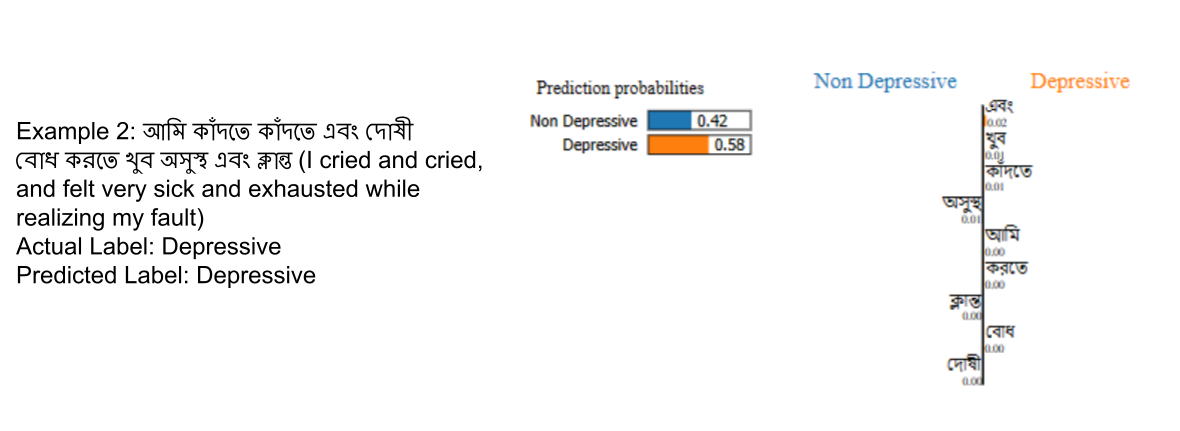} 
    \caption{A true negative sample example} 
    \label{fig:True_Negative}
\end{figure}

\begin{figure}[h!]
    \centering
    \includegraphics[width=0.9\textwidth]{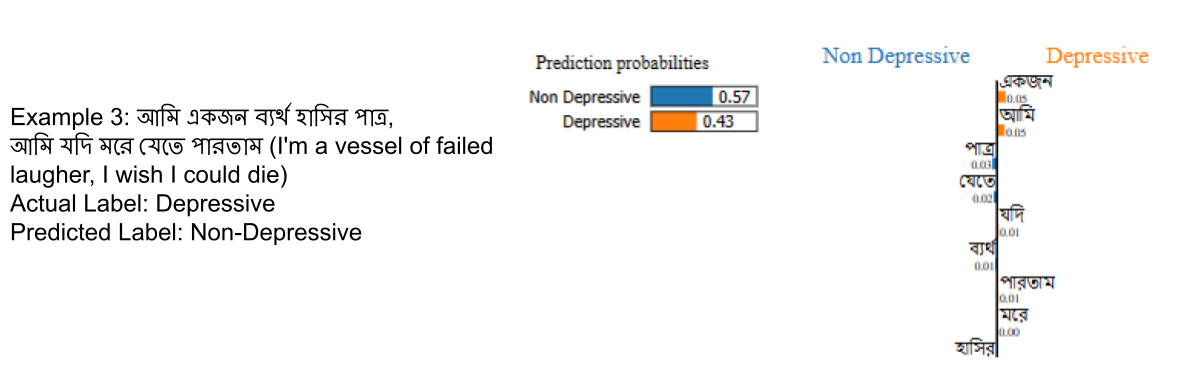} 
    \caption{A false negative sample example} 
    \label{fig:False_Negative}
\end{figure}

\begin{figure}[h!]
    \centering
    \includegraphics[width=0.9\textwidth]{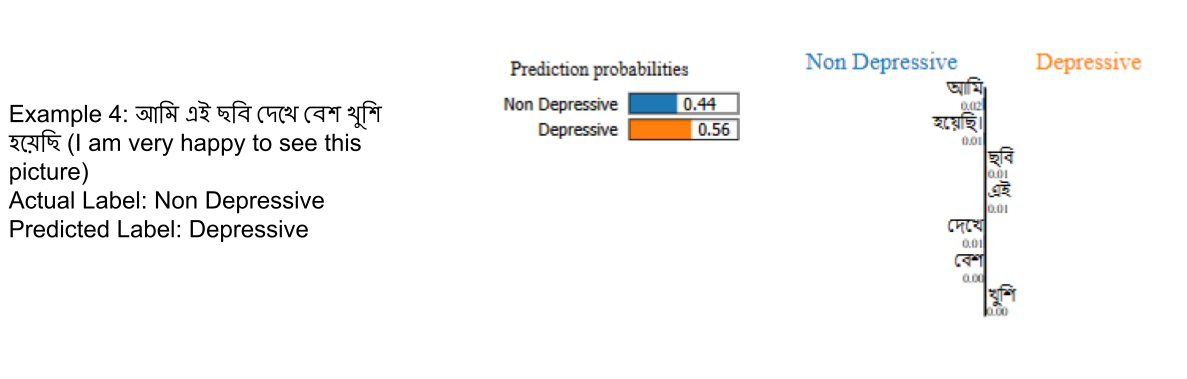} 
    \caption{A false positive sample example} 
    \label{fig:False_Positive}
\end{figure}

The study involved training transformer models for classifying text emotional content, using a dataset of 28,000 samples evenly split between depressive and non-depressive texts. Due to the nuanced nature of language, the models struggled with accurate classification. Explainability strategies were integrated to understand the decision-making process, focusing on word usage patterns and frequencies. The linguistic diversity of samples posed challenges in categorization, with distinct divisions into true positives (See Figure \ref{fig:True_Positive}) and true negatives(See Figure \ref{fig:True_Negative}). However, false negatives (depressive texts labelled as non-depressive) (See Figure \ref{fig:False_Negative}), and false positives (non-depressive texts labelled as depressive) (See Figure \ref{fig:False_Positive}) persisted, indicating the complexity of language. The implementation of explainability aimed to demystify model patterns, highlighting the need for more robust training sets to enhance precision in diverse linguistic contexts.

\subsection{Evaluation of Large Language Models}

Our proposed model DepGPT (GPT 3.5 Finetuned) outperforms the Alpaca Lora 7B in zero-shot and few-shot learning scenarios, with high recall and accuracy. GPT-4 has higher accuracy but lower recall. In few-shot scenarios, DepGPT has an F1-score of 0.9804, near-perfect recall, and exceptional precision. GPT-4 also performs admirably, earning excellent marks for all criteria. Despite improvements, GPT-3.5 Turbo and Alpaca Lora 7B perform less well overall.

The maximum performance of the Large Language Models is illustrated in the table \ref{tab:result-LLM}.
The full table of \ref{tab:result-LLM} is shown in \ref{tab:experiments-LLM-table2}.

\begin{table}[h!]
\centering
\caption{Performance of the Large Language Models}
\label{tab:result-LLM}
\begin{tabular}{c|c|cccc}
\hline
\multirow{2}{*}{\textbf{Category}} & \multirow{2}{*}{\textbf{Models}} & \multicolumn{4}{c}{\textbf{Performance Metrics}} \\ \cline{3-6} 
                           &                & Accuracy        & Precision & Recall & F1-score \\ \hline
\multirow{4}{*}{Zero-shot} & GPT-3.5 Turbo  & 0.8608          & 0.8931    & 0.8477 & 0.8698   \\
                           & DepGPT         & 0.9248          & 0.8899    & 0.9848 & 0.9349   \\
                           & GPT-4          & 0.8747          & 0.8452    & 0.9442 & 0.8921   \\
                           & Alpaca Lora 7B & 0.7549          & 0.7121    & 0.9291 & 0.8062   \\ \hline
\multirow{4}{*}{Few-shot}  & GPT-3.5 Turbo  & 0.8981          & 0.8846    & 0.9205 & 0.9022   \\
                           & DepGPT         & \textbf{0.9796} & 0.9615    & 0.9998 & 0.9804   \\
                           & GPT-4          & 0.9388          & 0.9231    & 0.9611 & 0.9412   \\
                           & Alpaca Lora 7B & 0.8571          & 0.8752    & 0.8409 & 0.8571   \\ \hline
\end{tabular}
\end{table}

\subsection{Comparison of all approaches}

The study compares various models used in our work, revealing that LLMs have superior accuracy compared to deep learning and transformer models. Transformer models have lower accuracy than deep learning models. The analysis reveals that our proposed model DepGPT (GPT-3.5 Turbo FineTuned) outperforms all models in adaptability and proficiency, excelling in zero-shot and few-shot tasks for large language models. Despite being competitive, GPT-3.5 Turbo and Alpaca Lora 7B show poorer efficacy across both learning scenarios.

\begin{figure}[h!]
    \centering
    \includegraphics[width=1.0\textwidth]{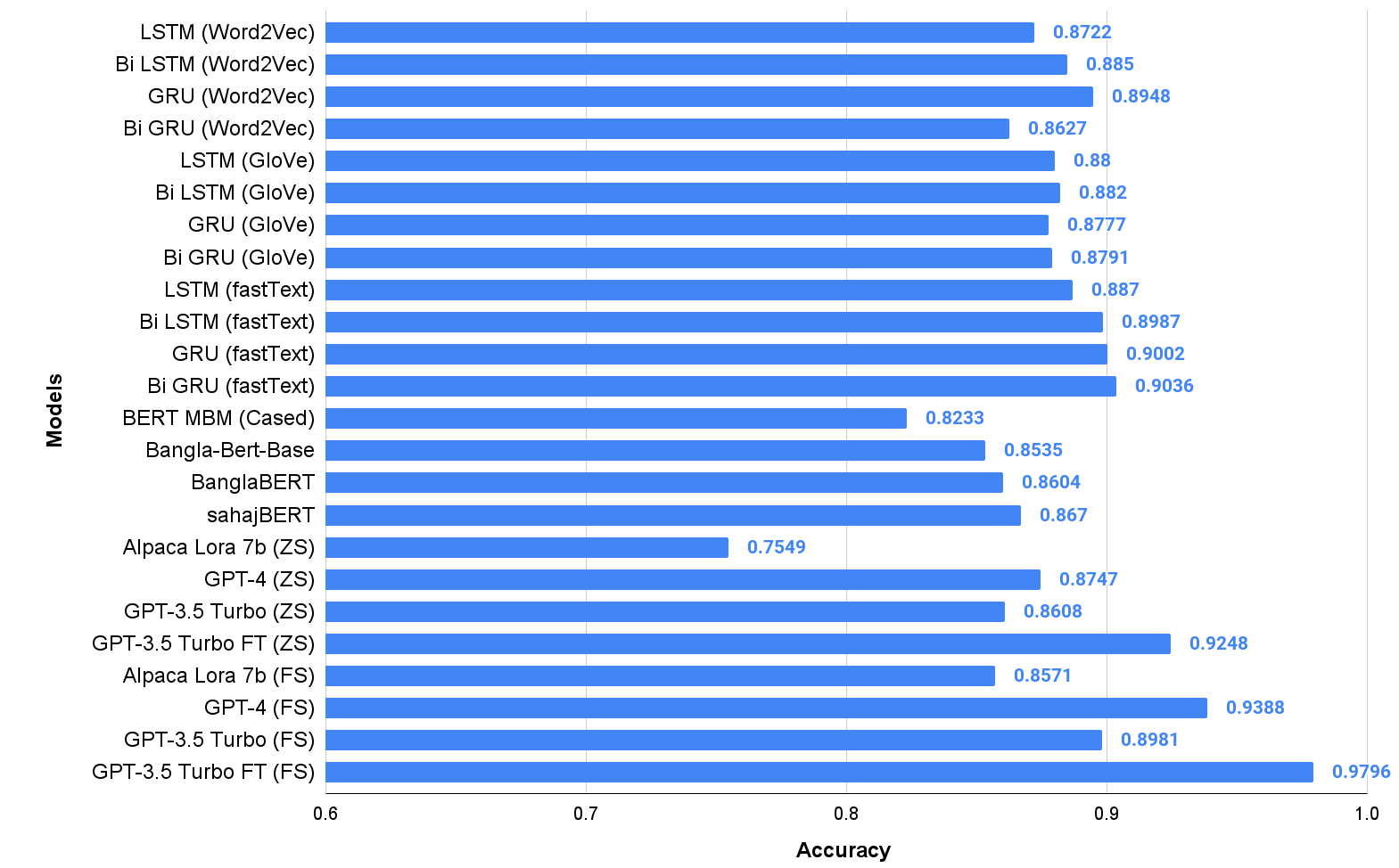} 
    \caption{Comparison of performance of all the models (Here MBM - BERT Multilingual Base (Cased), ZS - Zero shot, FS - Few shot, FT - Fine-tuned)} 
    \label{fig:all_result}
\end{figure}

\newpage

\section{Conclusion}

Deep learning models, transformer models, and large language models all have limits when it comes to sentiment analysis. These include data bias, context understanding, out-of-domain performance, language and dialect issues, long-term dependencies, resource reliance, interpretability, adaptability, generalization, dynamic language, emotional nuance, ethical considerations, and reliance on annotations. Transformer models understand context better, but they struggle with verbal nuances like sarcasm, irony, and inferred meanings. Large language models are computationally costly to train and fine-tune, limiting the capabilities of smaller groups of researchers. The dynamic nature of language evolution presents a risk since new idioms and slang can cause models to become out of date very fast. Complexity is increased when emotional nuances are expressed on a continuum as opposed to in binary terms. Two more important considerations in the assessment of these large-scale models are the ethical implications and the dependence on human annotations.

Future research on depression detection aims to enhance the resilience and sophistication of systems by developing adaptive models that can transition between domains and languages, detect subtle communicative signs, and respond in real time while maintaining ethical and user privacy. Researchers are also exploring continuous learning strategies and incorporating human input to improve accuracy in complex scenarios. The field is expanding to include interpreting emotions via text, music, and pictures, creating comprehensive benchmarking datasets, and researching unsupervised learning approaches. Improving security against malicious interventions is crucial for model dependability. Combining advanced sentiment analysis skills into online counselling platforms could revolutionize mental health support.

\bibliography{main}
\bibliographystyle{tmlr}

\appendix

\section{Result Tables of Deep Learning Models}

\begin{table}[h!]
\centering
\caption{Performance comparison of different sets of experiments in Deep Learning Architectures}
\label{tab:result-Deep-Learning2}
\resizebox{\columnwidth}{!}{%
\begin{tabular}{|c|c|c|c|c|c|c|c|c|c|c|c|}
\hline
\textbf{\begin{tabular}[c]{@{}c@{}}Model \\ Name\end{tabular}} &
  \textbf{Embedding} &
  \textbf{\begin{tabular}[c]{@{}c@{}}Batch \\ Size\end{tabular}} &
  \textbf{Epoch} &
  \textbf{\begin{tabular}[c]{@{}c@{}}Learning \\ Rate\end{tabular}} &
  \textbf{Dropout} &
  \textbf{\begin{tabular}[c]{@{}c@{}}Weght \\ Decay\end{tabular}} &
  \textbf{Layers} &
  \textbf{Accuracy} &
  \textbf{Precision} &
  \textbf{Recall} &
  \textbf{\begin{tabular}[c]{@{}c@{}}F1 \\ Score\end{tabular}} \\ \hline
\multirow{12}{*}{LSTM}   & \multirow{4}{*}{Glove}    & 16 & 10 & 0.001   & 0.2 & 0.001  & 2 & 0.88    & 0.88    & 0.88    & 0.88     \\ \cline{3-12} 
                         &                           & 16 & 20 & 0.001   & 0.4 & 0.01   & 3 & 0.864   & 0.866   & 0.864   & 0.863    \\ \cline{3-12} 
                         &                           & 32 & 15 & 0.001   & 0.2 & 0.001  & 3 & 0.859   & 0.86    & 0.859   & 0.858    \\ \cline{3-12} 
                         &                           & 32 & 25 & 0.001   & 0.3 & 0.005  & 4 & 0.8705  & 0.8706  & 0.87    & 8,705    \\ \cline{2-12} 
                         & \multirow{4}{*}{Word2Vec} & 16 & 15 & 0.001   & 0.3 & 0.001  & 3 & 0.8568  & 0.8589  & 0.8568  & 0.8566   \\ \cline{3-12} 
                         &                           & 32 & 20 & 0.001   & 0.2 & 0.001  & 3 & 0.8722  & 0.8748  & 0.8722  & 0.8719   \\ \cline{3-12} 
                         &                           & 32 & 25 & 0.001   & 0.4 & 0.01   & 2 & 0.8713  & 0.8753  & 0.8713  & 0.871    \\ \cline{3-12} 
                         &                           & 64 & 15 & 0.001   & 0.2 & 0.001  & 3 & 0.87193 & 0.87194 & 0.8719  & 0.8719   \\ \cline{2-12} 
                         & \multirow{4}{*}{fastText} & 16 & 20 & 0.01    & 0.2 & 0.005  & 2 & 0.8775  & 0.8782  & 0.8775  & 0.8774   \\ \cline{3-12} 
                         &                           & 16 & 15 & 0.001   & 0.2 & 0.0001 & 2 & 0.8814  & 0.8834  & 0.8814  & 0.88126  \\ \cline{3-12} 
                         &                           & 32 & 25 & 0.01    & 0.2 & 0.001  & 4 & 0.887   & 0.888   & 0.887   & 0.887    \\ \cline{3-12} 
                         &                           & 64 & 20 & 0.001   & 0.2 & 0.001  & 3 & 0.8828  & 0.884   & 0.8828  & 0.882719 \\ \hline
\multirow{13}{*}{BiLSTM} & \multirow{4}{*}{Glove}    & 16 & 10 & 0.001   & 0.3 & 0.001  & 2 & 0.882   & 0.88    & 0.88    & 0.882    \\ \cline{3-12} 
                         &                           & 32 & 20 & 0.001   & 0.2 & 0.005  & 3 & 0.862   & 0,86    & 0.862   & 0.862    \\ \cline{3-12} 
                         &                           & 64 & 20 & 0.001   & 0.4 & 0.001  & 2 & 0.866   & 0.87    & 0.866   & 0.865    \\ \cline{3-12} 
                         &                           & 64 & 25 & 0.001   & 0.3 & 0.01   & 4 & 0.8814  & 0.8816  & 0.8814  & 0.8813   \\ \cline{2-12} 
                         & \multirow{4}{*}{Word2Vec} & 16 & 15 & 0.001   & 0.3 & 0.001  & 4 & 0.8786  & 0.8793  & 0.8786  & 0.8785   \\ \cline{3-12} 
                         &                           & 32 & 20 & 0.001   & 0.2 & 0.005  & 3 & 0.885   & 0.8857  & 0.885   & 0.8849   \\ \cline{3-12} 
                         &                           & 32 & 25 & 0.001   & 0.3 & 0.01   & 2 & 0.8763  & 0.8768  & 0.8764  & 0.8763   \\ \cline{3-12} 
                         &                           & 64 & 30 & 0.001   & 0.2 & 0.001  & 3 & 0.8722  & 0.8735  & 0.8722  & 0.872    \\ \cline{2-12} 
                         & \multirow{5}{*}{fastText} & 16 & 20 & 0.001   & 0.2 & 0.005  & 2 & 0.8889  & 0.8889  & 0.88895 & 0.88894  \\ \cline{3-12} 
                         &                           & 32 & 20 & 0.001   & 0.2 & 0.01   & 2 & 0.886   & 0.8909  & 0.886   & 0.886    \\ \cline{3-12} 
                         &                           & 32 & 25 & 0.001   & 0.4 & 0.001  & 4 & 0.886   & 0.8903  & 0.886   & 0.885    \\ \cline{3-12} 
                         &                           & 64 & 20 & 0.001   & 0.2 & 0.001  & 3 & 0.8987  & 0.89874 & 0.8987  & 0.89871  \\ \cline{3-12} 
                         &                           & 64 & 30 & 0.00001 & 0.2 & 0.001  & 3 & 0.88895 & 0.8889  & 0.88895 & 0.88894  \\ \hline
\multirow{12}{*}{GRU}    & \multirow{4}{*}{Glove}    & 16 & 15 & 0.001   & 0.2 & 0.001  & 2 & 0.8716  & 0.8729  & 0.8716  & 0.8715   \\ \cline{3-12} 
                         &                           & 16 & 20 & 0.001   & 0.2 & 0.001  & 3 & 0.8777  & 0.878   & 0.8777  & 0.87777  \\ \cline{3-12} 
                         &                           & 32 & 25 & 0.001   & 0.2 & 0.001  & 3 & 0.8738  & 0.8748  & 0.8738  & 0.8738   \\ \cline{3-12} 
                         &                           & 64 & 30 & 0.001   & 0.3 & 0.001  & 3 & 0.8725  & 0.87408 & 0.8724  & 0.8723   \\ \cline{2-12} 
                         & \multirow{4}{*}{Word2Vec} & 16 & 15 & 0.001   & 0.3 & 0.001  & 2 & 0.8948  & 0.8959  & 0.8948  & 0.8947   \\ \cline{3-12} 
                         &                           & 32 & 20 & 0.001   & 0.2 & 0.005  & 4 & 0.8856  & 0.8862  & 0.8856  & 0.8855   \\ \cline{3-12} 
                         &                           & 32 & 25 & 0.001   & 0.4 & 0.01   & 2 & 0.8752  & 0.8753  & 0.8752  & 0.8752   \\ \cline{3-12} 
                         &                           & 64 & 30 & 0.001   & 0.3 & 0.001  & 3 & 0.8563  & 0.8594  & 0.8563  & 0.8559   \\ \cline{2-12} 
                         & \multirow{4}{*}{fastText} & 16 & 15 & 0.001   & 0.2 & 0.05   & 4 & 0.8984  & 0.8987  & 0.8984  & 0.8984   \\ \cline{3-12} 
                         &                           & 32 & 20 & 0.001   & 0.3 & 0.001  & 2 & 0.8891  & 0.8891  & 0.8891  & 0.8891   \\ \cline{3-12} 
                         &                           & 64 & 25 & 0.001   & 0.4 & 0.01   & 2 & 0.9002  & 0.9024  & 0.9002  & 0.9001   \\ \cline{3-12} 
                         &                           & 64 & 30 & 0.001   & 0.3 & 0.001  & 3 & 0.8882  & 0.8901  & 0.8882  & 0.888    \\ \hline
\multirow{12}{*}{BiGRU}  & \multirow{4}{*}{Glove}    & 16 & 15 & 0.001   & 0.2 & 0.001  & 2 & 0.8791  & 0.8834  & 0.8792  & 0.8788   \\ \cline{3-12} 
                         &                           & 32 & 20 & 0.001   & 0.3 & 0.001  & 3 & 0.8513  & 0.8517  & 0.8512  & 0.8512   \\ \cline{3-12} 
                         &                           & 32 & 25 & 0.001   & 0.3 & 0.001  & 2 & 0.8769  & 0.8784  & 0.8769  & 0.8768   \\ \cline{3-12} 
                         &                           & 64 & 30 & 0.001   & 0.2 & 0.001  & 2 & 0.847   & 0.8479  & 0.84709 & 0.847    \\ \cline{2-12} 
                         & \multirow{4}{*}{Word2Vec} & 16 & 15 & 0.001   & 0.3 & 0.001  & 3 & 0.8627  & 0.8636  & 0.8627  & 0.8626   \\ \cline{3-12} 
                         &                           & 32 & 20 & 0.001   & 0.2 & 0.001  & 3 & 0.8574  & 0.8595  & 0.8574  & 0.8572   \\ \cline{3-12} 
                         &                           & 32 & 25 & 0.001   & 0.4 & 0.01   & 2 & 0.8574  & 0.8575  & 0.8574  & 0.8574   \\ \cline{3-12} 
                         &                           & 64 & 30 & 0.001   & 0.2 & 0.001  & 3 & 0.8513  & 0.8554  & 0.8512  & 0.8508   \\ \cline{2-12} 
                         & \multirow{4}{*}{fastText} & 16 & 15 & 0.001   & 0.3 & 0.005  & 4 & 0.9036  & 0.9037  & 0.9036  & 0.9036   \\ \cline{3-12} 
                         &                           & 32 & 20 & 0.001   & 0.2 & 0.01   & 3 & 0.8991  & 0.8991  & 0.8991  & 0.8991   \\ \cline{3-12} 
                         &                           & 32 & 25 & 0.001   & 0.4 & 0.001  & 2 & 0.887   & 0.8952  & 0.887   & 0.8865   \\ \cline{3-12} 
                         &                           & 64 & 30 & 0.001   & 0.2 & 0.001  & 2 & 0.8965  & 0.8967  & 0.8965  & 0.8965   \\ \hline
\end{tabular}%
}
\end{table}

\newpage

\section{Result Tables of Large-scale Pre-trained Transformer Models (PLMs)}

\begin{table}[h!]
\centering
\caption{Performance of BERT Multilingual Base (Cased)}
\label{tab:result-bert-multi-table2}
\resizebox{\columnwidth}{!}{%
\begin{tabular}{|c|c|c|c|c|c|c|c|c|c|}
\hline
\textbf{Batch Size} &
  \textbf{Learning Rate} &
  \textbf{Num of epoch} &
  \textbf{Num of Folds} &
  \textbf{Momentum} &
  \textbf{Dropout} &
  \textbf{Accuracy} &
  \textbf{Precision} &
  \textbf{Recall} &
  \textbf{F1 Score} \\ \hline
\multirow{5}{*}{8} &
  \multirow{5}{*}{0.01} &
  10 &
  \multirow{31}{*}{5} &
  \multirow{31}{*}{0.9} &
  \multirow{31}{*}{0.1} &
  0.799 &
  0.755 &
  0.869 &
  0.8079 \\ \cline{3-3} \cline{7-10} 
                     &                        & 20 &  &  &  & 0.8111 & 0.7897 & 0.8481 & 0.8178 \\ \cline{3-3} \cline{7-10} 
                     &                        & 30 &  &  &  & 0.8136 & 0.791  & 0.8526 & 0.8206 \\ \cline{3-3} \cline{7-10} 
                     &                        & 40 &  &  &  & 0.8216 & 0.7794 & 0.8971 & 0.8341 \\ \cline{3-3} \cline{7-10} 
                     &                        & 50 &  &  &  & 0.8163 & 0.7935 & 0.8476 & 0.8196 \\ \cline{1-3} \cline{7-10} 
\multirow{4}{*}{16}  & \multirow{3}{*}{0.01}  & 10 &  &  &  & 0.8153 & 0.7738 & 0.8605 & 0.8149 \\ \cline{3-3} \cline{7-10} 
                     &                        & 20 &  &  &  & 0.8088 & 0.7618 & 0.8986 & 0.8245 \\ \cline{3-3} \cline{7-10} 
                     &                        & 30 &  &  &  & 0.8198 & 0.7792 & 0.8924 & 0.832  \\ \cline{2-3} \cline{7-10} 
                     & 0.001                  & 20 &  &  &  & 0.7886 & 0.7507 & 0.8642 & 0.8544 \\ \cline{1-3} \cline{7-10} 
\multirow{17}{*}{32} & 0.1                    & 50 &  &  &  & 0.8149 & 0.7923 & 0.8534 & 0.8217 \\ \cline{2-3} \cline{7-10} 
                     & 0.09                   & 20 &  &  &  & 0.8031 & 0.7791 & 0.8462 & 0.8113 \\ \cline{2-3} \cline{7-10} 
                     & 0.07                   & 50 &  &  &  & 0.8181 & 0.7813 & 0.8834 & 0.8292 \\ \cline{2-3} \cline{7-10} 
                     & 0.05                   & 50 &  &  &  & 0.8199 & 0.7792 & 0.8928 & 0.8321 \\ \cline{2-3} \cline{7-10} 
                     & 0.03                   & 50 &  &  &  & 0.8233 & 0.7821 & 0.8964 & 0.8354 \\ \cline{2-3} \cline{7-10} 
                     & \multirow{5}{*}{0.01}  & 10 &  &  &  & 0.7954 & 0.7652 & 0.8523 & 0.8064 \\ \cline{3-3} \cline{7-10} 
                     &                        & 20 &  &  &  & 0.8155 & 0.7722 & 0.8952 & 0.8291 \\ \cline{3-3} \cline{7-10} 
                     &                        & 30 &  &  &  & 0.8109 & 0.7729 & 0.8806 & 0.8233 \\ \cline{3-3} \cline{7-10} 
                     &                        & 40 &  &  &  & 0.8201 & 0.7808 & 0.8901 & 0.8319 \\ \cline{3-3} \cline{7-10} 
                     &                        & 50 &  &  &  & 0.8206 & 0.7799 & 0.8934 & 0.8328 \\ \cline{2-3} \cline{7-10} 
                     & 0.007                  & 50 &  &  &  & 0.8174 & 0.7796 & 0.8851 & 0.829  \\ \cline{2-3} \cline{7-10} 
                     & \multirow{3}{*}{0.005} & 50 &  &  &  & 0.788  & 0.7555 & 0.8518 & 0.8007 \\ \cline{3-3} \cline{7-10} 
                     &                        & 50 &  &  &  & 0.7897 & 0.7572 & 0.8528 & 0.8022 \\ \cline{3-3} \cline{7-10} 
                     &                        & 50 &  &  &  & 0.8186 & 0.7814 & 0.8849 & 0.8299 \\ \cline{2-3} \cline{7-10} 
                     & 0.003                  & 50 &  &  &  & 0.8113 & 0.77   & 0.8877 & 0.8247 \\ \cline{2-3} \cline{7-10} 
                     & 0.001                  & 50 &  &  &  & 0.7934 & 0.7575 & 0.8631 & 0.8068 \\ \cline{2-3} \cline{7-10} 
                     & 0.0005                 & 50 &  &  &  & 0.7775 & 0.745  & 0.8437 & 0.7913 \\ \cline{1-3} \cline{7-10} 
\multirow{5}{*}{64}  & \multirow{5}{*}{0.01}  & 10 &  &  &  & 0.7996 & 0.7576 & 0.8812 & 0.8147 \\ \cline{3-3} \cline{7-10} 
                     &                        & 20 &  &  &  & 0.8049 & 0.7697 & 0.8702 & 0.8035 \\ \cline{3-3} \cline{7-10} 
                     &                        & 30 &  &  &  & 0.802  & 0.763  & 0.8761 & 0.8156 \\ \cline{3-3} \cline{7-10} 
                     &                        & 40 &  &  &  & 0.8201 & 0.7858 & 0.8802 & 0.8303 \\ \cline{3-3} \cline{7-10} 
                     &                        & 50 &  &  &  & 0.8189 & 0.7785 & 0.8452 & 0.8104 \\ \hline
\end{tabular}%
}
\end{table}

\begin{table}[h!]
\centering
\caption{Performance of BanglaBERT}
\label{tab:result-BanglaBERT-table2}
\resizebox{\columnwidth}{!}{%
\begin{tabular}{|c|c|c|c|c|c|cccc|}
\hline
\textbf{Batch Size} &
  \textbf{Learning Rate} &
  \textbf{Num of epoch} &
  \textbf{Num of Folds} &
  \textbf{Momentum} &
  \textbf{Dropout} &
  \textbf{Accuracy} &
  \textbf{Precision} &
  \textbf{Recall} &
  \textbf{F1 Score} \\ \hline
\multirow{5}{*}{8} &
  \multirow{24}{*}{0.01} &
  10 &
  \multirow{24}{*}{5} &
  \multirow{24}{*}{0.9} &
  \multirow{24}{*}{0.1} &
  0.8443 &
  \textbf{0.8692} &
  0.8106 &
  0.8389 \\
                     &  & 20 &  &  &  & 0.8456          & 0.8607 & 0.8246          & 0.8423          \\
                     &  & 30 &  &  &  & 0.8338          & 0.8382 & 0.8274          & 0.8328          \\
                     &  & 40 &  &  &  & 0.8362          & 0.8304 & 0.845           & 0.8376          \\
                     &  & 50 &  &  &  & 0.8411          & 0.8397 & 0.8433          & 0.8415          \\ \cline{1-1} \cline{3-3} \cline{7-10} 
\multirow{4}{*}{16}  &  & 10 &  &  &  & 0.8474          & 0.8227 & 0.8858          & 0.853           \\
                     &  & 20 &  &  &  & 0.8381          & 0.846  & 0.8266          & 0.8362          \\
                     &  & 30 &  &  &  & 0.8538          & 0.8342 & 0.8831          & 0.8579          \\
                     &  & 50 &  &  &  & 0.853           & 0.8673 & 0.8335          & 0.8501          \\ \cline{1-1} \cline{3-3} \cline{7-10} 
\multirow{5}{*}{32}  &  & 10 &  &  &  & 0.8381          & 0.8094 & 0.8844          & 0.8453          \\
                     &  & 20 &  &  &  & 0.8494          & 0.8373 & 0.8673          & 0.8521          \\
                     &  & 30 &  &  &  & 0.8567          & 0.8325 & \textbf{0.8932} & 0.8618          \\
                     &  & 40 &  &  &  & 0.8551          & 0.8426 & 0.8734          & 0.8577          \\
                     &  & 50 &  &  &  & 0.8599          & 0.8435 & 0.8714          & 0.8572          \\ \cline{1-1} \cline{3-3} \cline{7-10} 
\multirow{5}{*}{64}  &  & 10 &  &  &  & 0.8515          & 0.8367 & 0.8735          & 0.8547          \\
                     &  & 20 &  &  &  & \textbf{0.8604} & 0.8575 & 0.8646          & 0.861           \\
                     &  & 30 &  &  &  & 0.8593          & 0.8477 & 0.8761          & 0.8617          \\
                     &  & 40 &  &  &  & 0.8554          & 0.8618 & 0.8465          & 0.8541          \\
                     &  & 50 &  &  &  & 0.8546          & 0.85   & 0.8613          & 0.8556          \\ \cline{1-1} \cline{3-3} \cline{7-10} 
\multirow{5}{*}{128} &  & 10 &  &  &  & 0.8458          & 0.8276 & 0.8736          & 0.85            \\
                     &  & 20 &  &  &  & 0.8504          & 0.842  & 0.8626          & 0.8522          \\
                     &  & 30 &  &  &  & 0.86            & 0.8472 & 0.8784          & \textbf{0.8625} \\
                     &  & 40 &  &  &  & 0.8574          & 0.8493 & 0.8688          & 0.859           \\
                     &  & 50 &  &  &  & 0.858           & 0.8514 & 0.8673          & 0.8593          \\ \hline
\end{tabular}%
}
\end{table}

\begin{table}[h!]
\centering
\caption{Performance of sahajBERT}
\label{tab:result-sahajBERT-table2}
\resizebox{\columnwidth}{!}{%
\begin{tabular}{|c|c|c|c|c|c|cccc|}
\hline
Batch Size &
  Learning Rate &
  Num of epoch &
  Num of Folds &
  Momentum &
  Dropout &
  Accuracy &
  Precision &
  Recall &
  F1 Score \\ \hline
\multirow{4}{*}{8} &
  \multirow{20}{*}{0.01} &
  10 &
  \multirow{20}{*}{5} &
  \multirow{20}{*}{0.9} &
  \multirow{20}{*}{0.1} &
  0.7171 &
  0.7175 &
  0.7162 &
  0.7169 \\
                     &  & 15 &  &  &  & 0.7357 & 0.7337 & 0.74   & 0.7368 \\
                     &  & 17 &  &  &  & 0.7205 & 0.7078 & 0.751  & 0.7288 \\
                     &  & 20 &  &  &  & 0.7335 & 0.7332 & 0.7362 & 0.7342 \\ \cline{1-1} \cline{3-3} \cline{7-10} 
\multirow{4}{*}{16}  &  & 7  &  &  &  & 0.7229 & 0.7172 & 0.7359 & 0.7264 \\
                     &  & 10 &  &  &  & 0.7411 & 0.7471 & 0.729  & 0.7379 \\
                     &  & 13 &  &  &  & 0.7508 & 0.7401 & 0.7729 & 0.7562 \\
                     &  & 17 &  &  &  & 0.756  & 0.7619 & 0.7448 & 0.7532 \\ \cline{1-1} \cline{3-3} \cline{7-10} 
\multirow{4}{*}{32}  &  & 7  &  &  &  & 0.7451 & 0.7417 & 0.752  & 0.7468 \\
                     &  & 10 &  &  &  & 0.7405 & 0.7567 & 0.7089 & 0.732  \\
                     &  & 13 &  &  &  & 0.7548 & 0.7524 & 0.7594 & 0.7559 \\
                     &  & 17 &  &  &  & 0.7422 & 0.7419 & 0.7429 & 0.7424 \\ \cline{1-1} \cline{3-3} \cline{7-10} 
\multirow{4}{*}{64}  &  & 7  &  &  &  & 0.7342 & 0.7345 & 0.7337 & 0.7341 \\
                     &  & 10 &  &  &  & 0.7285 & 0.7296 & 0.7261 & 0.7278 \\
                     &  & 13 &  &  &  & 0.7249 & 0.7286 & 0.7167 & 0.7226 \\
                     &  & 17 &  &  &  & 0.7539 & 0.7555 & 0.7509 & 0.7532 \\ \cline{1-1} \cline{3-3} \cline{7-10} 
\multirow{4}{*}{128} &  & 7  &  &  &  & 0.8506 & 0.8405 & 0.8655 & 0.8528 \\
                     &  & 10 &  &  &  & 0.863  & 0.85   & 0.8816 & 0.8655 \\
                     &  & 13 &  &  &  & 0.867  & 0.8718 & 0.8606 & 0.8662 \\
                     &  & 17 &  &  &  & 0.8618 & 0.8595 & 0.865  & 0.8622 \\ \hline
\end{tabular}%
}
\end{table}

\begin{table}[h!]
\centering
\caption{Performance of Bangla-BERT-Base}
\label{tab:result-Bangla-BERT-Base-table2}
\resizebox{\columnwidth}{!}{%
\begin{tabular}{|c|c|c|c|c|c|cccc|}
\hline
Batch Size &
  Learning Rate &
  Num of epoch &
  Num of Folds &
  Momentum &
  Dropout &
  \multicolumn{1}{c|}{Accuracy} &
  \multicolumn{1}{c|}{Precision} &
  \multicolumn{1}{c|}{Recall} &
  F1 Score \\ \hline
\multirow{5}{*}{8} &
  \multirow{25}{*}{0.01} &
  10 &
  \multirow{25}{*}{5} &
  \multirow{25}{*}{0.9} &
  \multirow{25}{*}{0.1} &
  0.7933 &
  0.7492 &
  0.8819 &
  0.8101 \\
                     &  & 20 &  &  &  & 0.8177 & 0.7914 & 0.8626 & 0.8255 \\
                     &  & 30 &  &  &  & 0.8326 & 0.8254 & 0.8436 & 0.8344 \\
                     &  & 40 &  &  &  & 0.8063 & 0.816  & 0.791  & 0.8033 \\
                     &  & 50 &  &  &  & 0.8051 & 0.7896 & 0.8319 & 0.8102 \\ \cline{1-1} \cline{3-3} \cline{7-10} 
\multirow{5}{*}{16}  &  & 10 &  &  &  & 0.842  & 0.8083 & 0.8967 & 0.8502 \\
                     &  & 20 &  &  &  & 0.8372 & 0.8227 & 0.8596 & 0.8407 \\
                     &  & 30 &  &  &  & 0.7994 & 0.7824 & 0.8295 & 0.8053 \\
                     &  & 40 &  &  &  & 0.8046 & 0.8061 & 0.8022 & 0.8042 \\
                     &  & 50 &  &  &  & 0.8011 & 0.788  & 0.824  & 0.8056 \\ \cline{1-1} \cline{3-3} \cline{7-10} 
\multirow{5}{*}{32}  &  & 10 &  &  &  & 0.8449 & 0.8405 & 0.8514 & 0.8459 \\
                     &  & 20 &  &  &  & 0.8437 & 0.8212 & 0.8787 & 0.849  \\
                     &  & 30 &  &  &  & 0.8042 & 0.7886 & 0.8313 & 0.8094 \\
                     &  & 40 &  &  &  & 0.825  & 0.7951 & 0.8757 & 0.8334 \\
                     &  & 50 &  &  &  & 0.832  & 0.8093 & 0.8687 & 0.838  \\ \cline{1-1} \cline{3-3} \cline{7-10} 
\multirow{5}{*}{64}  &  & 10 &  &  &  & 0.837  & 0.803  & 0.8931 & 0.8456 \\
                     &  & 20 &  &  &  & 0.8493 & 0.8362 & 0.8689 & 0.8522 \\
                     &  & 30 &  &  &  & 0.8356 & 0.8221 & 0.8565 & 0.8389 \\
                     &  & 40 &  &  &  & 0.832  & 0.8217 & 0.8479 & 0.8346 \\
                     &  & 50 &  &  &  & 0.824  & 0.815  & 0.8383 & 0.8265 \\ \cline{1-1} \cline{3-3} \cline{7-10} 
\multirow{5}{*}{128} &  & 10 &  &  &  & 0.8462 & 0.8373 & 0.8595 & 0.8482 \\
                     &  & 20 &  &  &  & 0.8417 & 0.8173 & 0.8803 & 0.8476 \\
                     &  & 30 &  &  &  & 0.8469 & 0.8243 & 0.8816 & 0.852  \\
                     &  & 40 &  &  &  & 0.8535 & 0.8315 & 0.8866 & 0.8582 \\
                     &  & 50 &  &  &  & 0.8503 & 0.8219 & 0.8943 & 0.8566 \\ \hline
\end{tabular}%
}
\end{table}

\newpage
\newpage
\section{Result Tables of Large Language Models}

\begin{table}[h!]
\centering
\caption{Performance comparison of different sets of experiments in LLM (where D - Depressive, ND - Non Depressive)}
\label{tab:experiments-LLM-table2}
\begin{tabular}{|c|c|cccc|c|}
\hline
\multirow{2}{*}{Category} &
  \multirow{2}{*}{Models} &
  \multicolumn{4}{c|}{Performance Metrics} &
  \multirow{2}{*}{Class} \\ \cline{3-6}
 &
   &
  \multicolumn{1}{c|}{Accuracy} &
  \multicolumn{1}{c|}{Precision} &
  \multicolumn{1}{c|}{Recall} &
  F1-score &
   \\ \hline
\multirow{8}{*}{Zero-shot Example} &
  \multirow{2}{*}{GPT-3.5 Turbo} &
  \multicolumn{1}{c|}{\multirow{2}{*}{0.8608}} &
  \multicolumn{1}{c|}{0.8931} &
  \multicolumn{1}{c|}{0.8477} &
  0.8698 &
  D \\ \cline{4-7} 
 &
   &
  \multicolumn{1}{c|}{} &
  \multicolumn{1}{c|}{0.8256} &
  \multicolumn{1}{c|}{0.8765} &
  0.8503 &
  ND \\ \cline{2-7} 
 &
  \multirow{2}{*}{DepGPT} &
  \multicolumn{1}{c|}{\multirow{2}{*}{0.9248}} &
  \multicolumn{1}{c|}{0.8899} &
  \multicolumn{1}{c|}{0.9848} &
  0.9349 &
  D \\ \cline{4-7} 
 &
   &
  \multicolumn{1}{c|}{} &
  \multicolumn{1}{c|}{0.9787} &
  \multicolumn{1}{c|}{0.8518} &
  0.9109 &
  ND \\ \cline{2-7} 
 &
  \multirow{2}{*}{GPT-4} &
  \multicolumn{1}{c|}{\multirow{2}{*}{0.8747}} &
  \multicolumn{1}{c|}{0.8452} &
  \multicolumn{1}{c|}{0.9442} &
  0.8921 &
  D \\ \cline{4-7} 
 &
   &
  \multicolumn{1}{c|}{} &
  \multicolumn{1}{c|}{0.9207} &
  \multicolumn{1}{c|}{0.7901} &
  0.8505 &
  ND \\ \cline{2-7} 
 &
  \multirow{2}{*}{Alpaca Lora 7B} &
  \multicolumn{1}{c|}{\multirow{2}{*}{0.7549}} &
  \multicolumn{1}{c|}{0.7121} &
  \multicolumn{1}{c|}{0.9291} &
  0.8062 &
  D \\ \cline{4-7} 
 &
   &
  \multicolumn{1}{c|}{} &
  \multicolumn{1}{c|}{0.8628} &
  \multicolumn{1}{c|}{0.5432} &
  0.6667 &
  ND \\ \hline
\multirow{8}{*}{Few-shot Example} &
  \multirow{2}{*}{GPT-3.5 Turbo} &
  \multicolumn{1}{c|}{\multirow{2}{*}{0.8981}} &
  \multicolumn{1}{c|}{0.8846} &
  \multicolumn{1}{c|}{0.9205} &
  0.9022 &
  D \\ \cline{4-7} 
 &
   &
  \multicolumn{1}{c|}{} &
  \multicolumn{1}{c|}{0.9131} &
  \multicolumn{1}{c|}{0.8751} &
  0.8936 &
  ND \\ \cline{2-7} 
 &
  \multirow{2}{*}{DepGPT} &
  \multicolumn{1}{c|}{\multirow{2}{*}{0.9796}} &
  \multicolumn{1}{c|}{0.9615} &
  \multicolumn{1}{c|}{0.9998} &
  0.9804 &
  D \\ \cline{4-7} 
 &
   &
  \multicolumn{1}{c|}{} &
  \multicolumn{1}{c|}{0.9981} &
  \multicolumn{1}{c|}{0.9583} &
  0.9787 &
  ND \\ \cline{2-7} 
 &
  \multirow{2}{*}{GPT-4} &
  \multicolumn{1}{c|}{\multirow{2}{*}{0.9388}} &
  \multicolumn{1}{c|}{0.9231} &
  \multicolumn{1}{c|}{0.9611} &
  0.9412 &
  D \\ \cline{4-7} 
 &
   &
  \multicolumn{1}{c|}{} &
  \multicolumn{1}{c|}{0.9565} &
  \multicolumn{1}{c|}{0.9167} &
  0.9362 &
  ND \\ \cline{2-7} 
 &
  \multirow{2}{*}{Alpaca Lora 7B} &
  \multicolumn{1}{c|}{\multirow{2}{*}{0.8571}} &
  \multicolumn{1}{c|}{0.8752} &
  \multicolumn{1}{c|}{0.8409} &
  0.8571 &
  D \\ \cline{4-7} 
 &
   &
  \multicolumn{1}{c|}{} &
  \multicolumn{1}{c|}{0.8401} &
  \multicolumn{1}{c|}{0.8751} &
  0.8571 &
  ND \\ \hline
\end{tabular}
\end{table}

\end{document}